\documentclass[a4paper,fleqn]{cas-dc}

\def\degree{${}^{\circ}$}

\usepackage[numbers]{natbib}
\usepackage{longtable}

\usepackage{booktabs}
\usepackage{appendix}
\def\tsc#1{\csdef{#1}{\textsc{\lowercase{#1}}\xspace}}
\tsc{CNN}
\tsc{Transformers}
\tsc{EP}
\tsc{PMS}
\tsc{BEC}
\tsc{DE}

\begin{document}

\shortauthors{Peng Zhao et~al.}
\let\printorcid\relax
\title [mode = title]{A Comparative Study of Deep Learning Classification Methods on a Small Environmental Microorganism Image Dataset (EMDS-6): from Convolutional Neural Networks to Visual Transformers}

\author[a]{Peng Zhao}

\author[a]{Chen Li}
\cormark[1]
\ead{lichen201096@hotmail.com}

\author[a]{Md Mamunur Rahaman}

\author[a]{Hao Xu}

\author[a]{Hechen Yang}

\author[d]{Hongzan Sun}

\author[b]{Tao Jiang}[style=chinese]

\author[c]{Marcin Grzegorzek}

\address[a]{Microscopic Image and Medical Image Analysis Group, MBIE College, Northeastern University, 110169, Shenyang, PR China}

\address[b]{School of Control Engineering, Chengdu 
University of Information Technology, Chengdu 610225, China}

\address[c]{University of Lübeck, Germany}

\address[d]{China Medical University, China}

\cortext[cor1]{Corresponding author}

\begin{abstract}
\
In recent years, deep learning has made brilliant achievements in \emph{Environmental Microorganism} (EM) image classification. However, image classification of small EM datasets has still not obtained good research results.  Therefore, researchers need to spend a lot of time searching for models with good classification performance and suitable for the current equipment working environment. To provide reliable references for researchers, we conduct a series of comparison experiments on 21 deep learning models. The experiment includes direct classification, imbalanced training, and hyperparameter tuning experiments. During the experiments, we find complementarities among the 21 models, which is the basis for feature fusion related experiments. We also find that the data augmentation method of geometric deformation is difficult to improve the performance of VTs (ViT, DeiT, BotNet and T2T-ViT) series models. In terms of model performance, Xception has the best classification performance, the ViT model consumes the least time for training, and the ShuffleNet-V2 model has the least number of parameters.

\end{abstract}


\maketitle

\section{Introduction}

With the advancement of industrialization, industrial pollution becomes increasingly serious. Therefore, finding effective methods to control, reduce or eliminate pollution is a top priority. Biological approaches have outstanding performance in solving environmental pollution problems. The Biological approaches have four main advantages in environmental treatment: no new pollution, no additional energy consumption, gentle process, decomposition products can feedback to nature and make a virtuous cycle of material changes~\cite{mckinney_2004_environmental}. Microorganisms are all tiny creatures that are invisible to the naked eyes. They are tiny and simple in structure, and usually can only be seen with a microscope. \emph{Environmental microorganisms} (EMs) specifically refer to those species of microorganisms that live in natural environments (such as mountains, streams and oceans) and artificial environments (such orchards and fish ponds). EMs play a vital role in whole nature for better or worse. For example, lactic acid bacteria can decompose some organic matter in the natural environment to provide nutrients for plants; actinomycetes can digest organic waste in sludge and improve water quality; microalgae can fix carbon dioxide in the air and be used as a raw material for biodiesel~\cite{zhao_2021_enhancement}; activated sludge composed of microorganisms has a strong ability to adsorb and oxidize organic matter and purify water~\cite{asgharnejad_2020_development}. harmful rhizosphere bacteria can inhibit plant growth by producing phytotoxins~\cite{fried_2000_monitoring}; Sludge bulking is caused by bacterial proliferation and the accumulation of sticky material, which poses a fundamental challenge for wastewater treatment~\cite{fan_2017_factors}. Therefore, EMs research helps solve environmental pollution problems, and the classification of EMs is the cornerstone of related research~\cite{emcuc_2018_Li}.





Generally, the size of EMs is between 0.1 and 100 $\mu m$, which is challenging to be identified and found. The traditional microbial classification method typically uses the " morphological method ", which requires a skilled operator to observe the EMs under a microscope. Then the results are given according to the shape characteristics. This is very time-consuming and financial~\cite{pepper_2011_environmental}. In addition, if researchers do not refer to the literature, even very experienced researchers cannot guarantee the accuracy and objectivity of the analysis results. Therefore, using the computer-aided classification of Environmental Microorganism (EM) images can enable researchers to use the slightest professional knowledge and the least time to make the most accurate judgments.

Currently, the analysis of EMs by computer vision is already achieved. For example, RGB (Red, Green, Blue) color analysis measures the number of microorganisms~\cite{sarrafzadeh_2015_microalgae, filzmoser_2011_review}, and deep learning methods are used to achieve the classification and segmentation of EM images. Among them, the research of EM classification using deep learning methods obtains more and more attention. Deep learning is a new research direction in the field of machine learning, and it provides good performance for image classification~\cite{zhang_2020_DLOGA}. Traditional machine learning-based EM classification methods rely on feature extraction, which requires many human resources ~\cite{ccayir_2018_FEBoD}. In contrast, deep learning-based algorithms perform feature extraction in an automated manner, allowing researchers to use minimal domain knowledge and workforce to extract prominent features. Furthermore, the classification results of deep learning are better than that of traditional machine learning in the case of super-large training samples~\cite{wang_2021_caoic}. However, for small datasets, the performance of deep learning is limited. Because the collection of EMs is usually carried out outdoors, for some sensitive EMs, transportation, storage and observation during the period may affect the quality of the final images. Therefore, it is difficult to obtain enough high-quality images, and this case results in the problem of small datasets. Therefore, this paper compares the performance of various deep learning models on small data sets of EMs and aims to find models with better performance on small data sets.



\begin{figure*}
	\centering
		\includegraphics[scale=0.51]{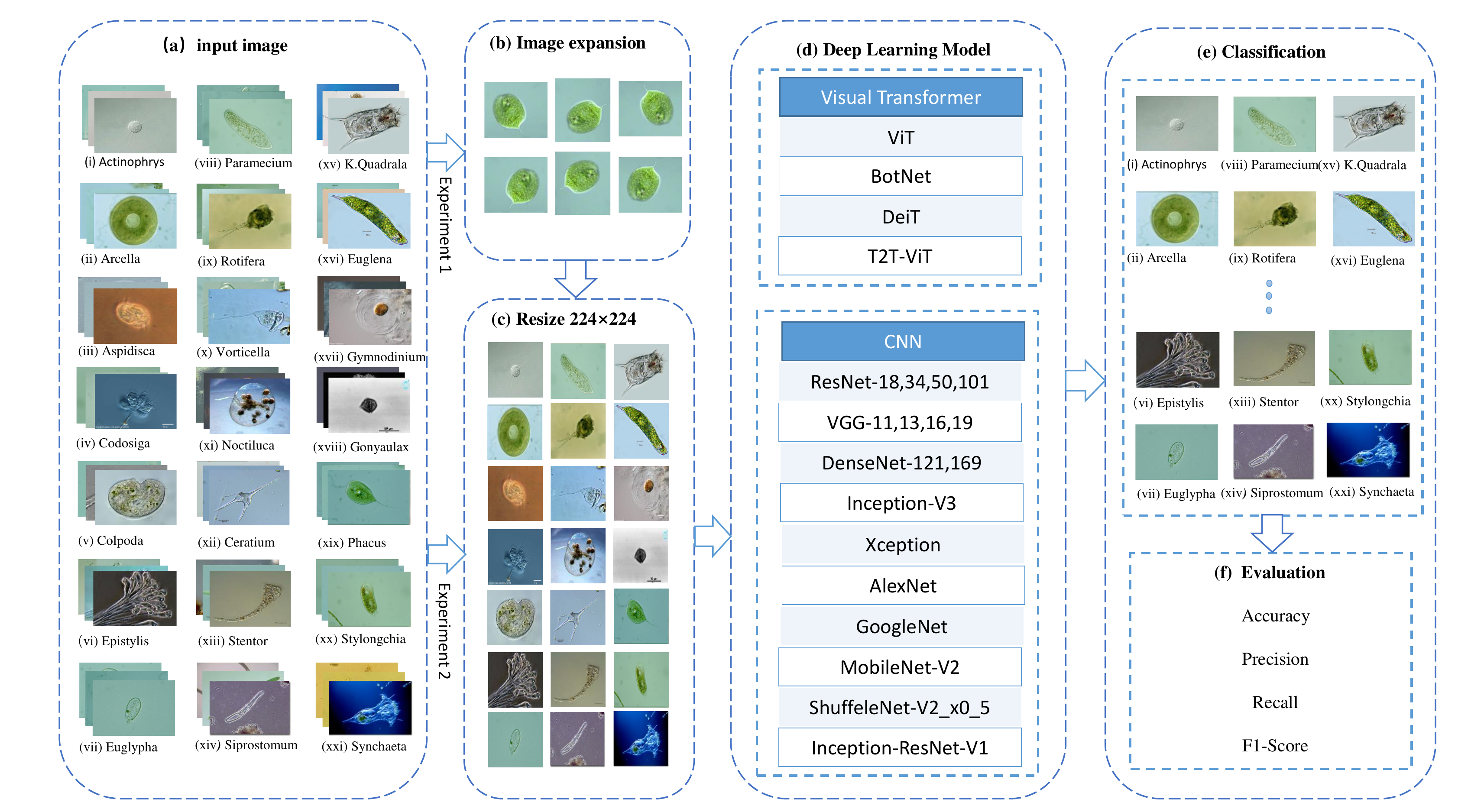}

	\caption{Environmental microbial classification process. (Vision Transformer (ViT), Bottleneck Transformers (BoTNet), Data-efficient image Transformers (DeiT), Tokens-to-Token Training Vision Transformer (T2T-ViT), Residual Network (ResNet), Visual Geometry Group (VGG))}
	\label{FIG:1}
\end{figure*}

This article compares a series of Convolutional Neural Networks (CNNs), such as ResNet-18, 34, 50, 101~\cite{he_2016_ResNet}, VGG11, 13, 16, 19~\cite{simonyan_2014_VGG}, DenseNet-121, 169~\cite{huang_2017_DenseNet}, Inception-V3~\cite{szegedy_2016_Inception-V3}, Xception~\cite{chollet_2017_xception}, AlexNet~\cite{krizhevsky_2012_AlexNet}, GoogleNet~\cite{szegedy_2015_GoogleNet}, MobileNet-V2~\cite{sandler_2018_mobilenetv2}, ShuffeleNet-V2x0.5~\cite{ma_2018_shufflenet}, Inception-ResNet-V1~\cite{szegedy_2017_inception-ResNet-v2}, and a series of visual transformers (VTs), such as vision transformer (ViT)~\cite{dosovitskiy_2020_ViT}, BotNet~\cite{srinivas_2021_BotNet}, DeiT~\cite{touvron_2020_Deit}, T2T-ViT~\cite{yuan_2021_T2T-ViT}. The purpose is to find deep learning models that are suitable for EM small datasets. The workflow diagram of this study is shown in Fig.\ref{FIG:1}. Step (b) is to rotate the training set and validation set images by 90\degree, 180\degree, 270\degree and mirror images up and down, left and right, augment the dataset by 6 times. Step (c) is uniform image size to 224×224 to facilitate training and classification. Step (d) is to input the processed data into different network models for training. Step (e) is to input the test set into the trained network for classification, and step (f) is to calculate the \emph{Average Precision} (AP), accuracy, precision, recall and F1-score based on the classification results to evaluate the performance of the network model. 

The structure of this paper is as follows. Section~\ref{sec:Related Work}  introduces the related methods of deep learning in image classification, the impact of small datasets on image classification, and the related work of deep learning models. Section~\ref{sec:Materials and Methods} introduces the dataset and experimental design in detail. Section~\ref{sec: Comparison of Classification Experiment} compares and summarizes the experimental results. Section~\ref{sec: Conclusion and Future Work} summarizes the whole paper and looks forward. 									

\section{Related Work\label{sec:Related Work}}

This section summarizes the impact of small datasets on classification, including basic deep learning image classification methods.

\subsection{The Impact of Small DataSets on Image Classification}

In rectal histopathology deep learning classification research, a large number of labeled pathological images are needed. However, the preparation of large datasets requires expensive labor costs and time costs, leading to the fact that existing studies are primarily based on small datasets. In addition, the lack of sufficient data leads to overfitting problems during the training process. A conditional sliding windows arithmetic is proposed in ~\cite{haryanto_2021_CSWAA} to solve this problem, which generates histopathological images. This arithmetic successfully solves the limitation of rectal histopathological data.

In climate research, the use of deep learning in cloud layer analysis often requires a lot of data. Therefore, classification in the case of a small dataset cannot achieve higher accuracy. In order to solve this problem,, a classification model with high accuracy on small datasets is proposed. The method improves from three aspects:

\noindent 1. A network model for a small dataset is designed.

\noindent 2. A regularization technique to increase the generalization ability of the model is applied.

\noindent 3. The average ensemble of models is used to improve the classification accuracy.

\noindent Therefore, the model not only has higher accuracy but also has better robustness~\cite{phung_2019_AHAMA}.

In deep learning research, small datasets often lead to classification over-fitting and low classification accuracy. According to this problem, a kind of deep CNN based transfer learning is designed to solve the problem of the small dataset. This method mainly improves data and models. In terms of data, the model transfers the feature layer of the CNN model pre-trained on big sample dataset to a small sample dataset. In terms of model, the whole series average pooling is used instead of the fully connected layer, and Softmax is used for classification. This method has a good classification performance on small sample datasets~\cite{zhao_2017_rotdl}.

Because of the limited training data, a two-phase classification method using migration learning and web data augmentation technology is proposed. This method increases the number of samples in the training set through network data augmentation. In addition, it reduces the requirements on the number of samples through transfer learning. This   classifier reduces the over-fitting problem while improving the generalization ability of the network ~\cite{han_2018_anicm}.

In radar image recognition, due to the complex environment and particular imaging principles, Synthetic Aperture Radar  (SAR) images have the problem of sample scarcity. A target recognition method of SAR image based on Constrained Naive Generative Adversarial Networks and CNN is proposed to solve this problem. This method combines Least Squares Generative Adversarial Networks and designs a shallow network structure based on the traditional CNNs model. The problem of high model complexity and over-fitting caused by the deep network structure is avoided, to improve the recognition performance. This method can better solve the problems of few image samples and intense speckle noise~\cite{mao_2021_Trosi}.

Lack of sufficient training data can seriously deteriorate the performance of neural networks and other classifiers. Due to this problem, a self-aware multi-classifier system suitable for "small data" cases is proposed. The system uses Neural Network, Support Vector Machines and Naive Bayes models as component classifiers. In addition, this system uses the confidence level as a criterion for classifier selection. The system performs well in various test cases and is incredibly accurate on small datasets~\cite{kholerdi_2018_eocos}.

CNNs are very effective for face recognition problems, but training such a network requires a large number of labeled images. Such large datasets are usually not public and challenging to collect. According to this situation, a method based on authentic face images to synthesize a vast training set is proposed. This method swaps the facial components of different face images to generate a new face. This technology achieves the most advanced face recognition performance on the Labled Faces in the Wild (LFW) face database~\cite{hu_2017_ldfru}.

The effectiveness of adjusting the number of convolutional layers to classify small datasets is proven in~\cite{chandrarathne_2020_AiEEp}. In addition, related experiments suggest that by employing a very low learning rate, the accuracy of classification of small datasets can be greatly increased.

In medical signal processing, very small datasets often lead to the problems of model overfitting and low classification accuracy. According to this situation, a method combining deep learning and traditional machine learning is proposed. This method uses the first few layers of CNN for feature extraction. Then, the extracted features are fed back to traditional supervised learning algorithms for classification. This method can avoid the overfitting problem caused by small datasets. In addition, it has better performance than traditional machine learning methods~\cite{alabandi_2017_CDLWT}.

\subsection{Deep Learning Technologies}

Due to the excellent performance of AlexNet in the image classification competition~\cite{krizhevsky_2012_AlexNet}, improvements in the CNN architecture are very active. A series of CNN-based networks continue to appear, making CNN an irreplaceable mainstream method in the field of computer vision. In recent years, Transformer frequently appears in computer vision tasks and provides good performance, which is sufficient to attract the attention of researchers.

\subsubsection{Convolutional Neural Networks}
AlexNet is the first large-scale CNN architecture to perform well in ImageNet classification. The innovation of the network lies in the successful application of the Rectified Linear Unit (Relu) activation function and the use of the Dropout mechanism and data enhancement strategy to prevent overfitting. To improve the model generalization ability, the network uses a Local Response Normalization layer. In addition, the maximum pooling of overlap is used to avoid the blurring effect caused by average pooling~\cite{krizhevsky_2012_AlexNet}.

The Visual Geometry Group of Oxford proposes the VGG network. The network uses a deeper network structure with depths of 11, 13, 16, and 19 layers. Meanwhile, VGG networks use a smaller convolution kernel (3×3 pixels) instead of the larger convolution kernel, which reduces the parameters and increases the expressive power of the networks~\cite{simonyan_2014_VGG}.

GoogLeNet is a deep neural network model based on the Inception module launched by Google. The network introduces an initial structure to increase the width and depth of the network while removing the fully connected layer and using average pooling instead of the fully connected layer to avoid the disappearance of the gradient. The network adds 2 additional softmax to conduct the gradient forward~\cite{szegedy_2015_GoogleNet}.

ResNet solves the "degradation" problem of deep neural networks by introducing residual structure. ResNet networks use multiple parameter layers to learn the representation of residuals between input and output, rather than using parameter layers to directly try to learn the mapping between input and output as VGGs networks do. Residual networks are characterized by ease of optimization and the ability to improve accuracy by adding considerable depth~\cite{he_2016_ResNet}.

The DenseNet network is inspired by the ResNet network. DenseNet uses a dense connection mechanism to connect all layers. This connection method allows the feature map learned by each layer to be directly transmitted to all subsequent layers as input, so that the features and the transmission of the gradient is more effective, and the network is easier to train. The network has the following advantages: it reduces the disappearance of gradients, strengthens the transfer of features, makes more effective use of features, and reduces the number of parameters to a certain extent~\cite{huang_2017_DenseNet}.

The inception-V3 network is mainly improved in two aspects. Firstly, branch structure is used to optimize the Inception Module; secondly, the larger two-dimensional convolution kernel is unpacked into two one-dimensional convolution kernels. This asymmetric structure can deal with more and richer spatial information and reduce the computation~\cite{szegedy_2016_Inception-V3}.

Xception is an improvement of Inception-V3. The network proposes a novel Depthwise Separable Convolution allign them in column, the core idea of which lies in space transformation and channel transformation. Compared with Inception, Xception has fewer parameters and is faster\cite{chollet_2017_xception}.

MobileNets and Xception have the same ideas but different pursuits. Xception pursues high precision, but MobileNets is a lightweight model, pursuing a balance between model compression and accuracy. A  new unit Inverted residual with linear bottleneck is applied in MobileNet-V2. The inverse residual first increases the number of channels, then performs convolution and then increases the number of channels. This can reduce memory consumption\cite{sandler_2018_mobilenetv2}.

ShuffleNet makes some improvements based on MobileNet. The 1×1 convolution used by MobileNet is a traditional convolution method with a lot of redundancy. However, ShuffleNet performs shuffle and group operations on 1×1 convolution. This operation implements channel shuffle and pointwise group convolution. In addition, this operation dramatically reduces the number of model calculations while maintaining accuracy~\cite{ma_2018_shufflenet}.

The Inception-ResNet network is inspired by ResNet, which introduces the residual structure of ResNet in the Inception module. Adding the residual structure does not significantly improve the model effect. But the residual structure helps to speed up the convergence and improve the calculation efficiency. The calculation amount of Inception-ResNet-v1 is the same as that of Inception-V3, but the convergence speed is faster~\cite{szegedy_2017_inception-ResNet-v2}.

\subsubsection{Visual Transformers}
The ViT model applies transformers in the field of natural language processing to the field of computer vision. The main contribution of this model is to prove that CNN is not the only choice for image classification tasks. ViT divides the input image into fixed-size patches and then obtains patch embedding through a linear transformation. Finally, the patch embeddings of the image are sent to the transformer to perform feature extraction to classification. The model is more effective than CNN on super-large-scale datasets and has high computational efficiency~\cite{dosovitskiy_2020_ViT}.

The BoTNet is proposed by Srinivas. This network introduces self-attention into ResNet. Therefore, BoTNet has both the local perception ability of CNN and the global information acquisition ability of Transformer. The top-1 accuracy on ImageNet is as high as 84.7\%, and the performance is better than models such as SENet and Efficient-Net~\cite{srinivas_2021_BotNet}.

T2T-ViT is an upgraded version of ViT. It proposes a novel Tokens-to-Token mechanism based on the characteristics and structure of ViT. This mechanism allows the deep learning model to model local and global information. The performance of this model is better than ResNet in the ImageNet data test, and the number of parameters and calculations are significantly reduced. In addition, the performance of its lightweight model is better than that of MobileNet~\cite{yuan_2021_T2T-ViT}.

DeiT is proposed by Touvron et al. The innovation of DeiT is proposes a new distillation process based on a distillation token, which has the same function as a class token. It is a token added after the image block sequence. The output after the transformer encoder and the output of the teacher model calculates the loss together. The training of DeiT requires fewer data and fewer computing resources~\cite{touvron_2020_Deit}.

\subsection{EM Image Classification\label{sec:EM Image Classification}}
With the development of technology, good results are achieved using computer-aided EM classification. In \cite{kruk_2015_computerized}, a system for automatic identification of different species of microorganisms in soil is proposed. The system first separates microorganisms from the background using the Otsu. Then shape features, edge features, and color histogram features are extracted. Then the features are filtered using a fast correlation-based filter. Finally, the random forest (RF) classifier is used for classification. This system frees researchers from the tedious task of microbial observation.

In \cite{amaral_1999_semi}, a semi-automatic microbial identification system is proposed. The system can accurately identify seven species of protozoa commonly found in wastewater. The system first enhances the image to be processed and then undergoes data collection and complex morphological operations to generate a 3D model of EMs. The 3D model is used to determine the species of protozoa. In \cite{amaral_2008_stalked}, a semi-automatic image analysis procedure is proposed. It is found that geometric features have good recognition ability. It is possible to detect the presence of two microorganisms, Opercularia and Vorticella, in wastewater plants. In \cite{cunshe_2008_new}, an improved neural network classification method based on microscopic images of sewage bacteria is proposed. The method uses principal component analysis to reduce the extracted EM features. Also, the method applies the daptivate accelerated back propagation (BP) algorithm to learn image classification. In \cite{xiaojuan_2009_improved}, an adaptive edge detection method is suggested for recognizing wastewater bacteria.  The method mentioned can effectively reduce image noise. Related experiments in the CECC database show that the proposed method enables rapid and accurate detection of wastewater bacteria.

An automatic classification method with high robustness of EMs is suggested in \cite{li_2013_classification}, which describes the shape of EMs in microscopic images by Edge Histograms, Extended Geometrical Features, etc. The support vector machine (SVM) classifier is used to achieve the best classification result of 89.7\%. A shape-based method for EM classification is suggested in \cite{yang_2014_shape}, which introduces very robust two-dimensional feature descriptors for EM shapes. The main process of this method is to separate EMs from the background. Then a new EM feature descriptor is used and finally a SVM is used for classification.

A new method for automatic classification of bacterial colony images is proposed in \cite{nie_2015_deep}, which enables the classification of colonies in different growth stages and contexts. In addition, the method mainly uses a multilayer middle layer CNN model for classification and uses the patches segmented from the CDBN model as input. Finally, a voting scheme is used for prediction. The results show that the method achieves results that exceed the classical model.

\section{Materials and Methods\label{sec:Materials and Methods}}

This section explains the EMDS-6 database, data augmentation methods, the distribution of the dataset, and the evaluation metrics for classification.

\subsection{Dataset}
\subsubsection{Data  Description}

\begin{figure*}
	\centering
		\includegraphics[scale=0.8]{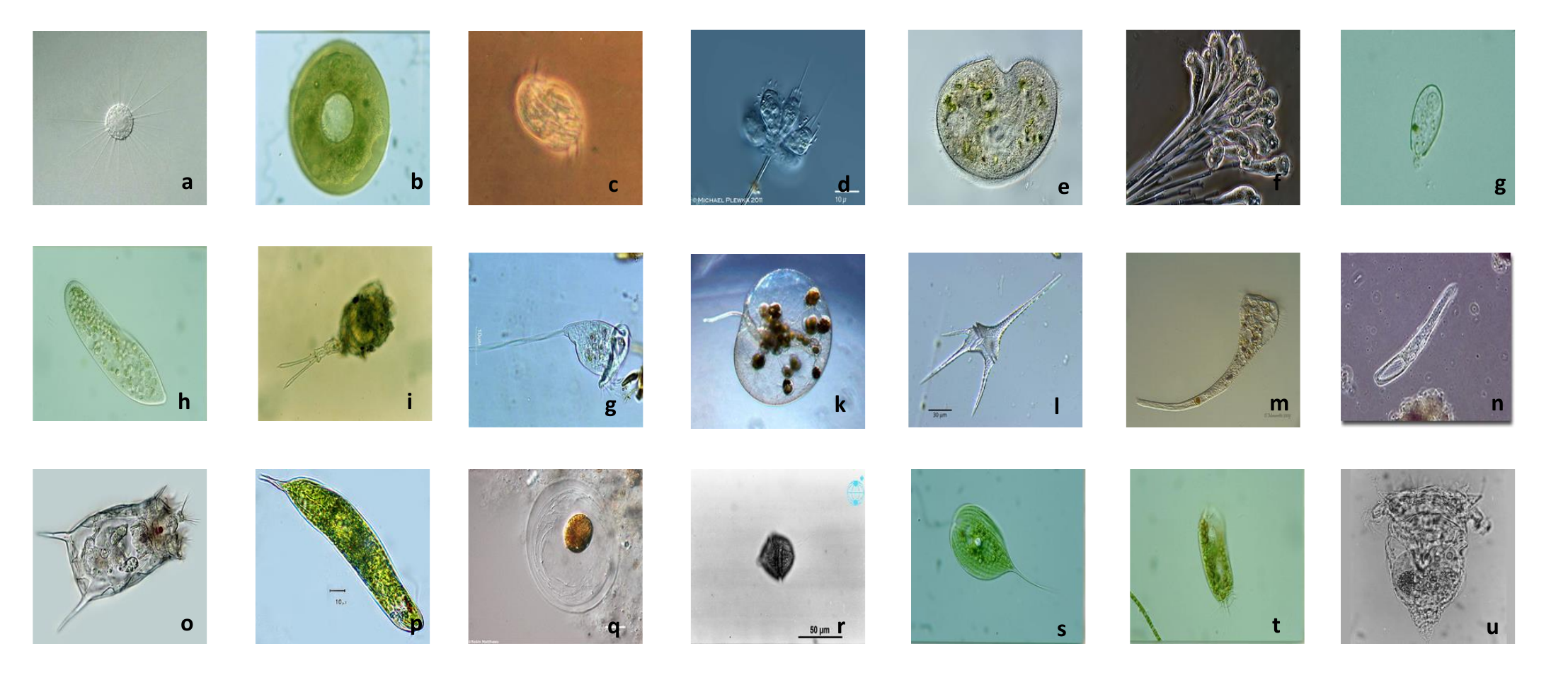}

	\caption{EMDS-6 database example:(a) \emph{Actinophrys}, (b) \emph{Arcella}, (c) \emph{Aspidisca}, (d) \emph{Codosiga}, (e) \emph{Colpoda}, (f) \emph{Epistylis}, (g) \emph{Euglypha}, (h) \emph{Paramecium}, (i) \emph{Rotifera}, (j) \emph{Vorticella}, (k) \emph{Noctiluca}, (l) \emph{Ceratium}, (m)  \emph{Stentor}, (n) \emph{Siprostomum}, (o) \emph{K. Quadrala}, (p) \emph{Euglena}, (q) \emph{Gymnodinium}, (r) \emph{Gymlyano}, (s) \emph{Phacus}, (t) \emph{Stylongchia}, (u) \emph{Synchaeta}. }
	\label{FIG:2}
\end{figure*}

This experiment uses Environmental Microorganism Dataset 6th Version (EMDS-6) to compare model performance. The database contains a total of 840 EM images of different sizes. These images contain a total of 21 types of environmental microorganisms, each with 40 images, namely: \emph{Actinophrys}, \emph{Arcella}, \emph{Aspidisca}, \emph{Codosiga}, \emph{Colpoda}, \emph{Epistylis}, \emph{Euglypha}, \emph{Paramecium}, \emph{Rotifera}, \emph{Vorticella}, \emph{Noctiluca}, \emph{Ceratium}, \emph{Stentor}, \emph{Siprostomum}, \emph{K. Quadrala}, \emph{Euglena}, \emph{Gymnodinium}, \emph{Gymlyano}, \emph{Phacus}, \emph{Stylongchia}, \emph{Synchaeta}. Some examples are shown in Fig. \ref{FIG:2}~\cite{li_2021_EMDS5}.

\subsubsection{Data Preprocessing}

In order to improve the accuracy of the model and reduce the degree of model overfitting, the images in EMDS-6 are augmented. Due to the security problem of data augmentation, the only geometric transformation of the data is performed here. The geometric transformation includes rotation 90\degree, 180\degree and 270\degree, up and down mirroring, and left and right mirroring. These transformations do not break the EM label and ensure data security. In addition, the image sizes in EMDS-6 is inconsistent, but the input required by the deep learning models is the same. Therefore, all images in EMDS-6 are standardized to 224×224 pixels.

\subsubsection{Data Settings \label{sec:Data settings}}

Experiment A: Randomly select 37.5\% of the database as the training set, 25\% as the validation set and 37.5\% as the test set. Experiment A is to directly perform classification tasks on 21 types of microorganisms through the deep learning model. The details of the training set, validation set, and test set are shown in Table. \ref{tbl1}. 

Experiment B: Randomly select 37.5\% of the database as the training set, 25\% as the validation set and 37.5\% as the test set. Specifically, 21 types of microorganisms are sequentially regarded as positive samples and the remaining 20 types of samples are regarded as negative samples. In this way, 21 new datasets are generated. For example, if \emph{Actinophrys} images are used as positive samples, the remaining 20 types of EMs such as \emph{Arcella} and {Aspidisca} are used as negative samples. Experiment B is imbalanced training to assist in verifying the performance of the model.

Because the EMDS-6 database is a very small dataset, the experimental results are quite contingent. Therefore, 37.5\% of the data is used to test the performance of the model to increase the reliability of the experiment. This also expresses our sincerity to the experimental results.

\begin{table}[11]
\caption{Database allocation details.}\label{tbl1}
\renewcommand\arraystretch{1}
\setlength{\tabcolsep}{4mm}{
\begin{tabular}{lccccc}
\toprule
Class\textbackslash{}Dataset & train & val & text & Total   \\
\midrule
\emph{Actinophrys}                  & 15    & 10  & 15   & 40      \\

\emph{Arcella}                      & 15    & 10  & 15   & 40     \\
\emph{Aspidisca}                    & 15    & 10  & 15   & 40      \\
\emph{Codosiga}                     & 15    & 10  & 15   & 40      \\
\emph{Colpoda}                      & 15    & 10  & 15   & 40      \\
\emph{Epistylis}                    & 15    & 10  & 15   & 40      \\
\emph{Euglypha}                     & 15    & 10  & 15   & 40      \\
\emph{Paramecium}                   & 15    & 10  & 15   & 40      \\
\emph{Rotifera}                     & 15    & 10  & 15   & 40      \\
\emph{Vorticella}                   & 15    & 10  & 15   & 40      \\
\emph{Noctiluca}                    & 15    & 10  & 15   & 40      \\
\emph{Ceratium}                     & 15    & 10  & 15   & 40      \\
\emph{Stentor}                      & 15    & 10  & 15   & 40      \\
\emph{Siprostomum}                  & 15    & 10  & 15   & 40      \\
\emph{K.Quadrala}                   & 15    & 10  & 15   & 40      \\
\emph{Euglena}                      & 15    & 10  & 15   & 40      \\
\emph{Gymnodinium}                  & 15    & 10  & 15   & 40      \\
\emph{Gonyaulax}                    & 15    & 10  & 15   & 40      \\
\emph{Phacus}                       & 15    & 10  & 15   & 40      \\
\emph{Stylongchia}                  & 15    & 10  & 15   & 40      \\
\emph{Synchaeta}                    & 15    & 10  & 15   & 40      \\
Total                        & 315   & 210 & 315  & 840     \\
\bottomrule
\end{tabular}}
\end{table}

\subsection{Evaluation Method}

To scientifically evaluate the classification performance of deep learning models, choosing appropriate indicators is a crucial factor. Recall, Precision, Accuracy, F1-score, \emph{Average Precision} (AP), and \emph{mean Average Precision} (mAP) are commonly used evaluation indicators~\cite{xie_2015_MICac}. The effectiveness of these indicators is proven. The Recall is the probability of being predicted to be positive in actual positive samples. Precision is the probability of being actual positive in all predicted positive samples. AP refers to the average value of recall rate from 0 to 1. The mAP is the arithmetic average of all AP.  F1-score is the harmonic value of precision rate and recall rate.  Accuracy refers to the percentage of correct results predicted in the total sample~\cite{powers_2020_EFRRA}.  The specific calculation methods of these indicators are shown in Table. \ref{tbl4}.

\begin{table}[]
\caption{Evaluation Metrics for Image Classification. Sample classification (K), Number of positive samples (M). }\label{tbl4}
\renewcommand\arraystretch{1.8}
\begin{tabular*}{\tblwidth}{@{} LL@{} }
\toprule
\makecell[c]{Assessments} & \makecell[c]{Formula} \\
\midrule
\makecell[c]{Precision ($P$)} & \makecell[c]{$\rm \frac{TP}{TP + FP}$} \\
\makecell[c]{Recall ($R$)} & \makecell[c]{$\rm \frac{TP}{TP + FN}$} \\

\makecell[c]{F1-score} & \makecell[c]{$ 2 \times \frac{P \times R}{P + R}$} \\
\makecell[c]{Accuracy} & \makecell[c]{$\rm \frac{TP+TN}{TP + TN + FP + FN}$} \\
\makecell[c]{AP} & \makecell[c]{$\rm \frac{1}{M} \sum_{i=1}^M Precision_{max}~(i)$}\\
\makecell[c]{mAP} & \makecell[c]{$\rm \frac{1}{K} \sum_{j=1}^K AP~(j)$}\\
\bottomrule
\end{tabular*}
\end{table}

In Table. \ref{tbl4}, TN is the number of negative classes predicted as negative classes, FP represents the number of negative classes predicted as positive classes, FN refers to the number of positive classes predicted as negative classes and TP is the number of positive classes predicted as positive classes.
\section{Comparison of Classification Experiment\label{sec: Comparison of Classification Experiment}}
\subsection{Experimental Environment}

This comparative experiment is performed on the local computer. The computer hardware configuration is shown in Table. \ref{tbl2}. The computer software configuration is as follows: Win10 Professional operating system, Python 3.6, and Pytorch 1.7.1. In addition, the code runs in the integrated development environment Pycharm 2020 Community Edition.

This experiment mainly uses some classic deep learning models and some relatively novel deep learning models. The hyperparameters uniformly set by these models are shown in Table. \ref{tbl5}.

\begin{table}[]
\caption{Computer hardware configuration.}\label{tbl2}
\renewcommand\arraystretch{1.8}
\begin{tabular*}{\tblwidth}{@{} LL@{} }
\toprule
Hardware  & Product Number \\
\midrule
CPU                    &    Intel Core i7-10700           \\
GPU                    &    NVIDIA Quadro RTX 4000       \\
Motherboard            &    HP 8750 (LPC Controller-0697)     \\
RAM                    &    SAMSUNG DDR4 3200MHz \\
SSD                    &    HP SSD S750 256GB        \\
\bottomrule
\end{tabular*}
\end{table}

\begin{table}[]
\caption{Deep learning model parameters}\label{tbl5}
\renewcommand\arraystretch{1.8}
\begin{tabular*}{\tblwidth}{@{} LL@{}}
\toprule
Parameter  & Value \\
\midrule
Batch Size & 32    \\
Epoch      & 100   \\
Learning   & 0.002 \\
Optimizer  & Adam  \\
\bottomrule
\end{tabular*}
\end{table}

\subsection{Experimental Results and Analysis}
\subsubsection{The Classification Performance of Each Model on the Training and Validation Sets}

\begin{figure*}
	\centering
		\includegraphics[scale=0.45]{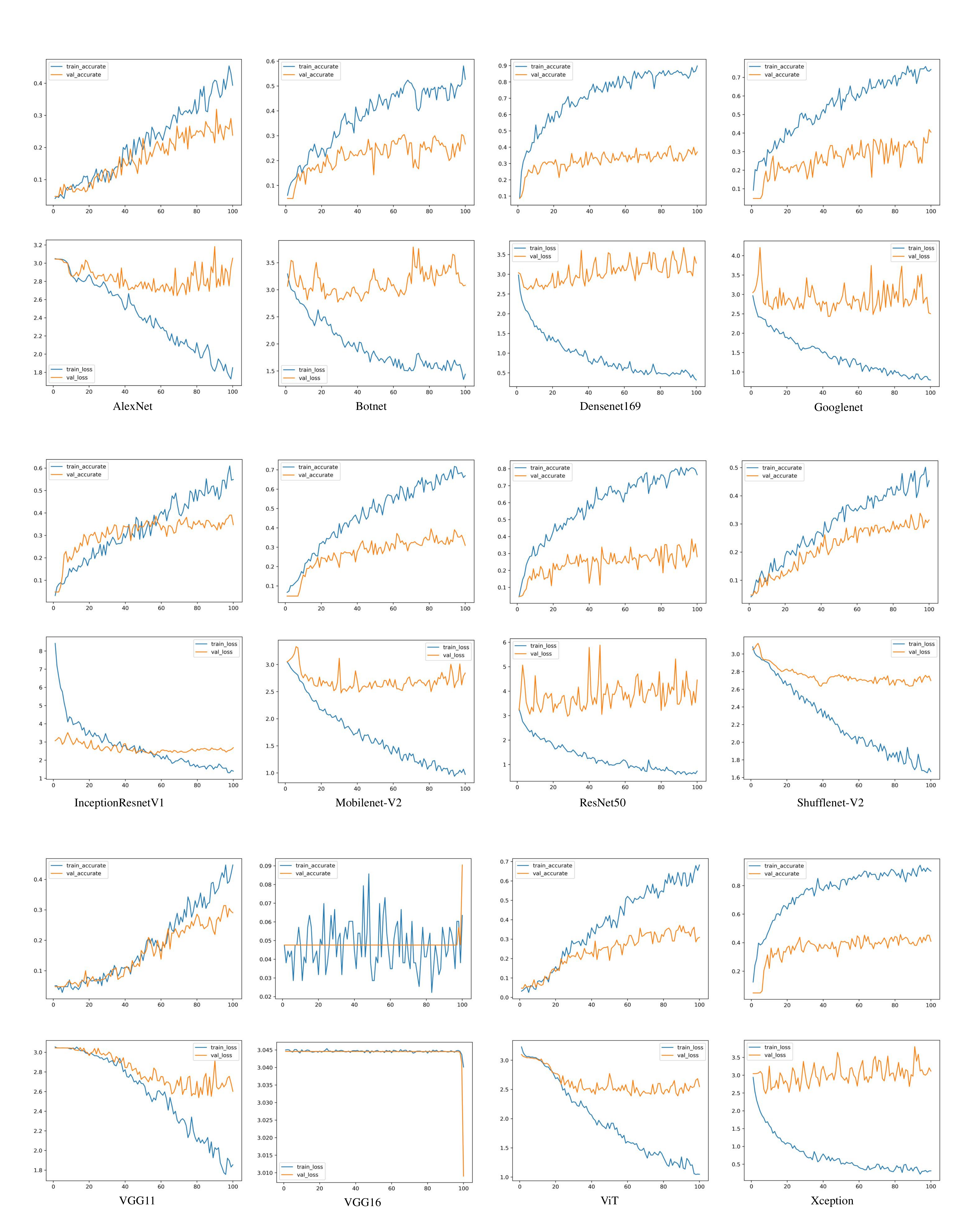}

	\caption{The loss curve and accuracy curve of different deep learning networks on the training set and validation set. For example, AlexNet, Botnet, Densenet169, Googlenet, InceptionResnet-V1, Mobilenet-V2, ResNet50, ShuffleNet-V2, VGG11, VGG16, ViT, and Xception.  train-accurate is the accuracy curve of the training set, train-accurate is the accuracy curve of the validation set, train-loss is the loss curve of the training set, and val-loss is the loss curve of the validation set.}

	\label{FIG:4}
\end{figure*}

Figure. \ref{FIG:4} shows the accuracy and loss curves of the CNNs and VT series models. Table. \ref{tbl6} shows the performance indicators of different deep learning models on the validation set. According to Figure. \ref{FIG:4} and Table. \ref{tbl6}, the performance of different deep learning models using small EM dataset cases is briefly evaluated.

As shown in Figure. \ref{FIG:4}, the accuracy rate of the training set is much higher than that of the validation set of each model. Densenet169, Googlenet, Mobilenet-V2, ResNet50, ViT, and Xception network models are particularly over-fitted. In addition, AlexNet, InceptionResnetV1, ShuffleNet-V2 and VGG11 network models do not show serious overfitting. Among 21 models in Table. \ref{tbl6}, the accuracy rates of the Deit, ViT and T2T-ViT models are at the 10th, 12th and 14th. The VT models are in the middle and downstream position among the 21 models.

The Xception network model has the highest accuracy, precision, and recall rates in the test set results, which are 40.32\%, 49.71\% and 40.33\%. The AlexNet, ViT, and ShuffleNet-V2 network models require the shortest training time, which are 711.64s, 714.56s and 712.95s. In addition, the ShuffleNet-V2 network model has the smallest parameter amount, which is 1.52MB.

VGG16 and VGG19 networks cannot converge in the EMDS-6 dataset classification task. The VGG13 network model has the lowest accuracy, precision and recall rates in the validation set results, which are 20.95\%, 19.23\% and 20.95\%. The VGG19 network model requires the longest training time, which is 1036.68s. In addition, the VGG19 network model has the largest amount of parameters, which is 521MB.

\begin{table*}[]
\renewcommand\arraystretch{1.3} 
\caption{After fusing the two features, it has ideal precision and ideal performance improvement. The left side of the table shows the improved accuracy of feature fusion under ideal conditions, and the right side of the table shows the accuracy of feature fusion under ideal conditions.}\label{bast}
\setlength{\tabcolsep}{4.5mm}{

\begin{tabular}{lllllllll}
\hline
\multicolumn{2}{l}{Model}             & Change (up) &  & \multicolumn{2}{l}{Model} & Accuracy \\ \hline
ResNet101         & VGG11         & 9.52\%   &  & Googlenet         & Xception & 46.03\%  \\
InceptionResnetV1 & ResNet18      & 7.94\%   &  & Inception-V3      & Xception & 44.76\%  \\
Inception-V3      & Shufflenet-V2 & 7.62\%   &  & ResNet50          & Xception & 44.76\%  \\
Shufflenet-V2     & VGG11         & 7.62\%   &  & Deit              & Xception & 44.44\%  \\
Deit              & VGG11         & 7.30\%   &  & Densenet161       & Xception & 44.13\%  \\
Inception-V3      & VGG11         & 7.30\%   &  & VGG11             & Xception & 44.13\%  \\
ResNet18          & ResNet50      & 7.30\%   &  & Densenet121       & Xception & 43.81\%  \\
ResNet34          & ResNet50      & 7.30\%   &  & Mobilenet-V2      & Xception & 43.81\%  \\
ResNet34          & VGG11         & 7.30\%   &  & ResNet34          & ResNet50 & 43.81\%  \\
ResNet101         & Shufflenet-V2 & 7.30\%   &  & ResNet34          & VGG11    & 43.81\%  \\
Googlenet         & Mobilenet-V2  & 7.30\%   &  & Densenet121       & ResNet34 & 43.49\%  \\
Alexnet           & T2T-ViT      & 6.98\%   &  & Googlenet         & ResNet34 & 43.49\%  \\
Deit              & Mobilenet-V2  & 6.98\%   &  & InceptionResnetV1 & Xception & 43.49\%  \\
Deit              & vit-5         & 6.98\%   &  & Mobilenet-V2      & ResNet34 & 43.49\%  \\
Densenet121       & Googlenet     & 6.98\%   &  & ResNet18          & Xception & 43.49\% \\ \hline
\end{tabular}}
\end{table*}
Xception is a network with excellent performance in the EMDS-6 database classification. In the Xception network accuracy curve, the accuracy of the Xception network training set is rising rapidly, approaching the highest point of 90\% after 80 epochs. Meanwhile, the accuracy of the validation set is close to the highest point 45\%, after 30 epochs. In addition, the Xception network training set loss curve declines steadily and approaches its lowest point after 80 epochs. But the validation set loss begins to approach the lowest point after 20 epochs and stops falling. VGG13 is a network that performs poorly on EMDS-6 classification. In the VGG13 network, the accuracy curve of the training set and the accuracy curve of the validation set have similar trends, and there are obvious differences after 80 epochs. Meanwhile, the loss of the training set and the loss of the validation set are also relatively close, and there are obvious differences after 60 epochs. Networks such as Xception, ResNet34 and Googlenet are relatively high-performance networks. The training accuracy of these networks is much higher than the validation accuracy. Furthermore, the validation accuracy is close to the highest point in a few epochs. In addition, the training set loss of these networks is usually lower than 0.3 at 100 epochs. VGG11 and AlexNet are poorly performing networks. These network training accuracy curves are relatively close to the validation accuracy curves. Disagreements usually occur after many epochs. In addition, the training set loss of these networks is usually higher than 0.3 at 100 epochs.

\begin{table*}
\caption{Comparison of classification results of different deep learning models on the \textbf{validation set}. P denotes Precision, and R represents Recall. (Sort in descending order of classification accuracy.) }\label{tbl6}
\renewcommand\arraystretch{0.55} 
\centering
\setlength{\tabcolsep}{3.2mm}{
\begin{tabular}{@{}ccccccccccc@{} }
\toprule

\multirow{2}{*}{Model}             & \multirow{2}{*}{Avg. R(\%)} & \multirow{2}{*}{Avg. P(\%)} & \multirow{2}{*}{Avg. F1\_score(\%)} & \multirow{2}{*}{Accuracy(\%)} & \multirow{2}{*}{Params Size (MB)} & \multirow{2}{*}{Time (s)} \\

                                   &                             &                                &                                &                               &                                   &                           \\ \midrule
\multirow{2}{*}{Xception}          & \multirow{2}{*}{45.71\%}    & \multirow{2}{*}{52.48\%}       & \multirow{2}{*}{44.95\%}       & \multirow{2}{*}{45.71\%}      & \multirow{2}{*}{79.8}             & \multirow{2}{*}{996}      \\
                                   &                             &                                &                                &                               &                                   &                           \\
\multirow{2}{*}{ResNet34}          & \multirow{2}{*}{42.86\%}    & \multirow{2}{*}{45.33\%}       & \multirow{2}{*}{42.31\%}       & \multirow{2}{*}{42.86\%}      & \multirow{2}{*}{81.3}             & \multirow{2}{*}{780}      \\
                                   &                             &                                &                                &                               &                                   &                           \\
\multirow{2}{*}{Googlenet}         & \multirow{2}{*}{41.90\%}    & \multirow{2}{*}{42.83\%}       & \multirow{2}{*}{40.49\%}       & \multirow{2}{*}{41.91\%}      & \multirow{2}{*}{21.6}             & \multirow{2}{*}{772}      \\
                                   &                             &                                &                                &                               &                                   &                           \\
\multirow{2}{*}{Densenet121}       & \multirow{2}{*}{40.95\%}    & \multirow{2}{*}{43.61\%}       & \multirow{2}{*}{40.09\%}       & \multirow{2}{*}{40.95\%}      & \multirow{2}{*}{27.1}             & \multirow{2}{*}{922}      \\
                                   &                             &                                &                                &                               &                                   &                           \\
\multirow{2}{*}{Densenet169}       & \multirow{2}{*}{40.95\%}    & \multirow{2}{*}{43.62\%}       & \multirow{2}{*}{39.89\%}       & \multirow{2}{*}{40.95\%}      & \multirow{2}{*}{48.7}             & \multirow{2}{*}{988}      \\
                                   &                             &                                &                                &                               &                                   &                           \\
\multirow{2}{*}{ResNet18}          & \multirow{2}{*}{40.95\%}    & \multirow{2}{*}{45.55\%}       & \multirow{2}{*}{41.05\%}       & \multirow{2}{*}{40.95\%}      & \multirow{2}{*}{42.7}             & \multirow{2}{*}{739}      \\
                                   &                             &                                &                                &                               &                                   &                           \\
\multirow{2}{*}{Inception-V3}      & \multirow{2}{*}{40.00\%}    & \multirow{2}{*}{45.01\%}       & \multirow{2}{*}{39.70\%}       & \multirow{2}{*}{40.00\%}      & \multirow{2}{*}{83.5}             & \multirow{2}{*}{892}      \\
                                   &                             &                                &                                &                               &                                   &                           \\
\multirow{2}{*}{Mobilenet-V2}      & \multirow{2}{*}{39.52\%}    & \multirow{2}{*}{39.57\%}       & \multirow{2}{*}{37.01\%}       & \multirow{2}{*}{39.52\%}      & \multirow{2}{*}{8.82}             & \multirow{2}{*}{767}      \\
                                   &                             &                                &                                &                               &                                   &                           \\
\multirow{2}{*}{InceptionResnetV1} & \multirow{2}{*}{39.05\%}    & \multirow{2}{*}{41.54\%}       & \multirow{2}{*}{37.96\%}       & \multirow{2}{*}{39.05\%}      & \multirow{2}{*}{30.9}             & \multirow{2}{*}{800}      \\
                                   &                             &                                &                                &                               &                                   &                           \\
\multirow{2}{*}{Deit}              & \multirow{2}{*}{39.05\%}    & \multirow{2}{*}{39.37\%}       & \multirow{2}{*}{37.70\%}       & \multirow{2}{*}{39.05\%}      & \multirow{2}{*}{21.1}           & \multirow{2}{*}{817.27}     \\
                                   &                             &                                &                                &                               &                                   &                           \\
\multirow{2}{*}{ResNet50}          & \multirow{2}{*}{38.57\%}    & \multirow{2}{*}{43.84\%}       & \multirow{2}{*}{38.02\%}       & \multirow{2}{*}{38.57\%}      & \multirow{2}{*}{90.1}             & \multirow{2}{*}{885}      \\
                                   &                             &                                &                                &                               &                                   &                           \\
\multirow{2}{*}{ViT}             & \multirow{2}{*}{37.14\%}    & \multirow{2}{*}{41.02\%}       & \multirow{2}{*}{35.95\%}       & \multirow{2}{*}{37.14\%}      & \multirow{2}{*}{31.2}             & \multirow{2}{*}{715}      \\
                                   &                             &                                &                                &                               &                                   &                           \\
\multirow{2}{*}{ResNet101}         & \multirow{2}{*}{34.76\%}    & \multirow{2}{*}{36.52\%}       & \multirow{2}{*}{32.99\%}       & \multirow{2}{*}{34.76\%}      & \multirow{2}{*}{162}              & \multirow{2}{*}{1021}     \\
                                   &                             &                                &                                &                               &                                   &                           \\
\multirow{2}{*}{T2T-ViT}           & \multirow{2}{*}{34.29\%}    & \multirow{2}{*}{38.17\%}       & \multirow{2}{*}{34.54\%}       & \multirow{2}{*}{34.28\%}      & \multirow{2}{*}{15.5}            & \multirow{2}{*}{825.3}     \\
                                   &                             &                                &                                &                               &                                   &                           \\
\multirow{2}{*}{ShuffleNet-V2}     & \multirow{2}{*}{33.81\%}    & \multirow{2}{*}{33.90\%}       & \multirow{2}{*}{31.68\%}       & \multirow{2}{*}{33.81\%}      & \multirow{2}{*}{1.52}             & \multirow{2}{*}{713}      \\
                                   &                             &                                &                                &                               &                                   &                           \\
\multirow{2}{*}{AlexNet}           & \multirow{2}{*}{31.90\%}    & \multirow{2}{*}{32.53\%}       & \multirow{2}{*}{29.32\%}       & \multirow{2}{*}{31.91\%}      & \multirow{2}{*}{217}              & \multirow{2}{*}{712}      \\
                                   &                             &                                &                                &                               &                                   &                           \\
\multirow{2}{*}{VGG11}             & \multirow{2}{*}{31.43\%}    & \multirow{2}{*}{41.20\%}       & \multirow{2}{*}{29.97\%}       & \multirow{2}{*}{31.43\%}      & \multirow{2}{*}{491}              & \multirow{2}{*}{864}      \\
                                   &                             &                                &                                &                               &                                   &                           \\
\multirow{2}{*}{BotNet}            & \multirow{2}{*}{30.48\%}    & \multirow{2}{*}{32.61\%}       & \multirow{2}{*}{30.06\%}       & \multirow{2}{*}{30.48\%}      & \multirow{2}{*}{72.2}             & \multirow{2}{*}{894}      \\
                                   &                             &                                &                                &                               &                                   &                           \\
\multirow{2}{*}{VGG13}             & \multirow{2}{*}{20.95\%}    & \multirow{2}{*}{19.23\%}       & \multirow{2}{*}{18.37\%}       & \multirow{2}{*}{20.95\%}      & \multirow{2}{*}{492}              & \multirow{2}{*}{957}      \\
                                   &                             &                                &                                &                               &                                   &                           \\
\multirow{2}{*}{VGG16}             & \multirow{2}{*}{9.05\%}     & \multirow{2}{*}{1.31\%}        & \multirow{2}{*}{2.10\%}        & \multirow{2}{*}{9.05\%}       & \multirow{2}{*}{512}              & \multirow{2}{*}{990}      \\
                                   &                             &                                &                                &                               &                                   &                           \\
\multirow{2}{*}{VGG19}             & \multirow{2}{*}{4.76\%}     & \multirow{2}{*}{0.23\%}        & \multirow{2}{*}{0.44\%}        & \multirow{2}{*}{4.76\%}       & \multirow{2}{*}{532}              & \multirow{2}{*}{1036}     \\
                                   &                             &                                &                                &                               &                                   &  \\                        

\bottomrule
\end{tabular}}
\end{table*}

\subsubsection{The Classification Performance of Each Model on Test Set}

\begin{figure*}
	\centering
		\includegraphics[scale=0.44]{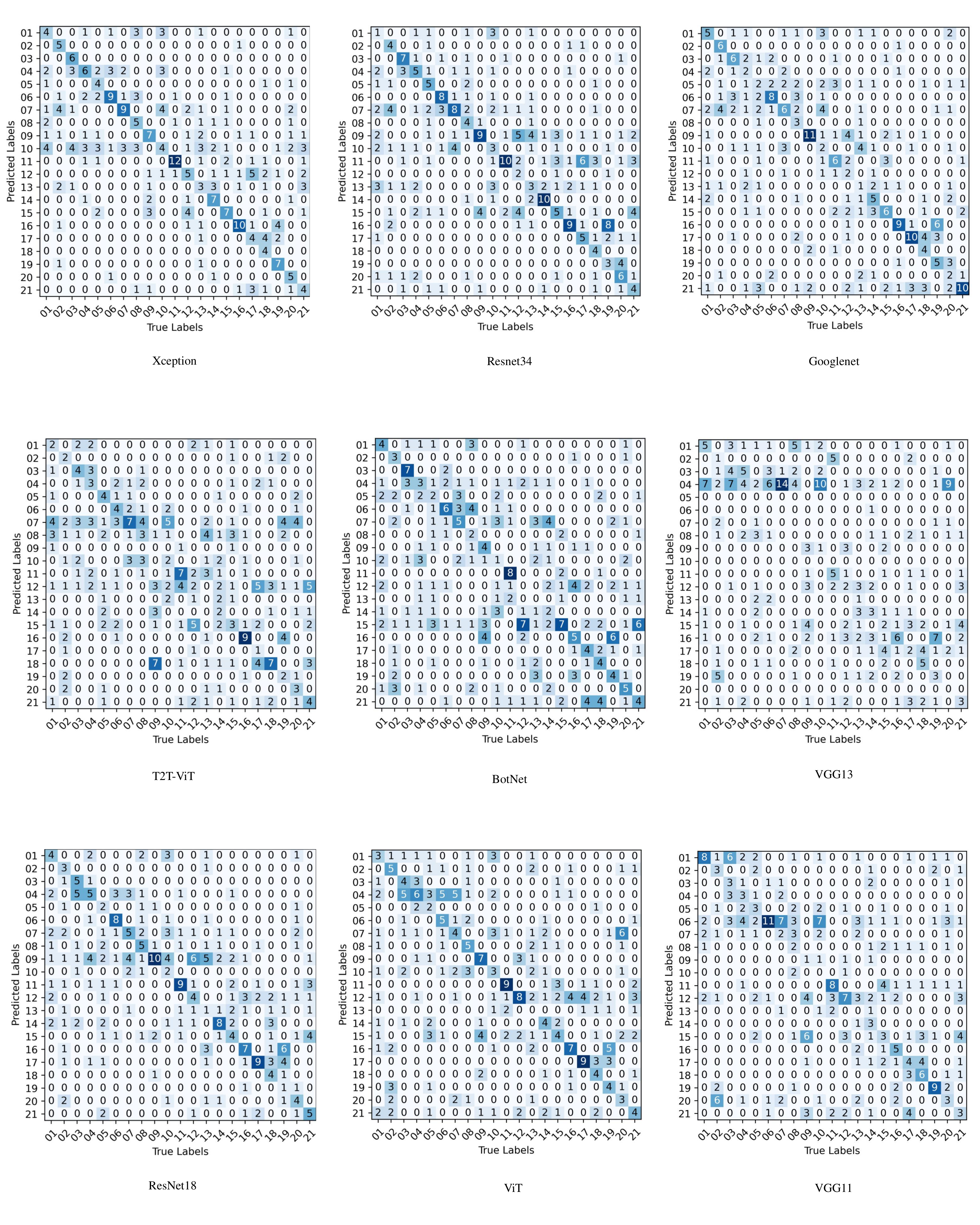}

	\caption{ Confusion matrix comparison of different network models on test set, Xception, Resnet34, Googlenet, T2T-ViT, BotNet, VGG13,ResNet18,ViT, and VGG11. ( In the confusion matrix, 01, 02, 03, 04, 05, 06, 07, 08, 09, 10, 11, 12, 13, 14, 15, 16, 17, 18, 19, 20, 21 represent Actinophrys, Arcella, Aspidisca, Codosiga, Colpoda, Epistylis, Euglypha, Paramecium, Rotifera, Vorticella, Noctiluca, Ceratium, Stentor, Siprostomum, K. Quadrala, Euglena, Gymnodinium, Gymlyano, Phacus, Stylongchia and Synchaeta, respectively. )}
	\label{FIG:5}
\end{figure*}

Table. \ref{tbl7}, shows the performance indicators of each model on the test set, including precision, recall, F1-score, and accuracy. Moreover, the confusion matrix of the CNNs and VTs models are shown in Figure. \ref{FIG:5}.

It is observed from the test set results that the accuracy ranking of each model remains unchanged. The accuracy rate of the Xception network on the test set is still ranked first, at 40.32\%, and is 3.81\% higher than the second. Meanwhile, the average accuracy, average recall rate and average F1-score of the Xception network also remain in the first place, at 40.32\%, 40.33\% and 41.41\%. Excluding the non-convergent VGG16 and VGG19 networks, the accuracy of the VGG13 validation set is still ranked at the bottom, at 15.55\%. However, the ranking of the T2T-ViT network on the validation set accuracy rate changes dramatically. The accuracy rate of the T2T-ViT network is 34.28\%, and the ranking rose from 12th to 5th. In addition, the average precision, average recall and average F1-score of the T2T-ViT network are 38.17\%, 34.29\% and 34.54\%. Judging from the time consumed for the models, the ViT model consumes the least time at 3.77s.  On the other hand, the Densenet169 model consumes the most time at 11.13s.

Figure. \ref{FIG:5} depicts the confusion matrix generated by part of the test dataset to more intuitively show the classification performance of the CNNs and VTs models on small EM datasets. In Table. \ref{tbl7}, Xception is the network with the best overall performance, and VGG13 is the network with the worst overall performance. In the confusion matrix of the Xception network, 127 EM images out of 315 EM images are classified into the correct category. In addition, the 11th type of EM classification performs the best, with 12 EM images are correctly classified and 3 EM images are misclassified into other categories. Meanwhile, the Xception network performs the worst in the 13th category of EM classification results. 3 EM images are correctly classified and 14 EM images are misclassified into other categories. For the VGG13 network, 49 of the 315 EM images are classified into the correct category. Among them, the 16th EM classification performs best. 6 EM images are correctly classified, and 9 EM images are mistakenly classified into other categories. Comparing the CNNs and VTs models, all of the models perform well on the 11th EM classification and perform poorly on the 13th EM classification. For example, the ViT model correctly classifies 9 EM images and 0 EM images in the classification of the 11th and 13th class EMs, respectively.	


Figure~\ref{FIG:5} shows that Xception better classifies the 11th and 16th types of EM images. ResNet is better at classifying tasks of the 11th and 16th types of images. Googlenet is better at classifying the 9th, 17th, and 21st EMs. The overall classification performance of T2T-ViT is poor. However, there are still outstanding performances in the 16th EM classification. The BotNet hybrid model is good at the 11th type of EM classification. However, the classification performance on the 12th and 13th images is abysmal. ResNet is good at image classification in the 9th, 11th, and 17th categories. The ViT model is good at the 11th, 12th, and 17th EM image classification. It is found from Figure~\ref{FIG:5} that the images that each model is good at classifying are not the same. Therefore, there is a certain degree of complementarity among different deep learning models.


From Figure \ref{FIG:5}, Xception and Googlenet are highly complementary. For example, Googlenet has a good performance in the classification of EMs in classes 17 and 21, but Xception has a poor performance in the classification of EMs in classes 17 and 21. In addition, Xception is better at classifying the 11th class of EM images than Googlenet. This result shows that the features extracted by the two models are quite different. Two networks can extract features that each other network cannot extract. Therefore, there is a strong complementarity between the two features. In addition, although VGG11 performs poorly in the classification of EMs. However, VGG11 is better at class 1 and class 19 classification tasks than Resnet34. Therefore, there is still a certain complementarity between the features extracted by the two models. This complementarity makes it possible to improve model performance through feature fusion.


In the study, we combine 18 models in pairs. Regardless of the specific feature fusion method or the possibility of a particular implementation, we calculate the ideal performance of the two models after fusion based on the current results. Part of the results is shown in Table~\ref{bast}. All results of the table are in the appendix. In table~\ref{bast}, the ideal accuracy rate of each combination is calculated by the following steps. For each combination, the best results of every model are firstly accumulated. Then, the accumulated results are divided by the total number of images in the test set, and the result is the ideal accuracy rate. For example, the combination of Xception and Googlenet. In class 1 EM classification, Xception correctly classifies 4 images, and Googlenet correctly classifies 5 images. Here, 5 are the best results. The other categories can be deduced by analogy. The calculation method of model performance improvement is as follows: Use the ideal accuracy to subtract the highest accuracy of the two models to obtain the performance that can be improved in the ideal state after the fusion. In Table~\ref{bast}, the fusion of Xception and Googlenet performs best on the EMDS-6, with a classification accuracy of 46.03\%. However, ResNet101 and VGG11 are improved the most after the fusion, and the two models have the strongest complementarity. On the left side of Table~\ref{bast}, we can clearly see the ideal effect of improving the accuracy after the fusion of the two features. The improved accuracy after fusion reflects the complementarity of the two models to some extent. This complementarity can provide some help to researchers who are engaged in feature fusion.

\begin{table*}
\caption{Comparison of classification results of different deep learning models in the test set. P denotes Precision, and R represents Recall. (Sort in descending order of classification accuracy.)}\label{tbl7}
\renewcommand\arraystretch{0.55} 
\centering
\setlength{\tabcolsep}{3.2mm}{
\begin{tabular}{@{}ccccccccccc@{} }
\toprule

\multirow{2}{*}{Model}             & \multirow{2}{*}{Avg. R(\%)} & \multirow{2}{*}{Avg. P(\%)} & \multirow{2}{*}{Avg. F1\_score(\%)} & \multirow{2}{*}{Accuracy(\%)} & \multirow{2}{*}{Params Size (MB)} & \multirow{2}{*}{Time (s)} \\
                                   &                             &                                &                                &                               &                                  &                          \\ \midrule
\multirow{2}{*}{Xception}          & \multirow{2}{*}{40.33\%}    & \multirow{2}{*}{49.71\%}       & \multirow{2}{*}{41.41\%}       & \multirow{2}{*}{40.32\%}      & \multirow{2}{*}{79.8}            & \multirow{2}{*}{5.63}    \\
                                   &                             &                                &                                &                               &                                  &                          \\
\multirow{2}{*}{ResNet34}          & \multirow{2}{*}{36.51\%}    & \multirow{2}{*}{42.92\%}       & \multirow{2}{*}{36.22\%}       & \multirow{2}{*}{36.51\%}      & \multirow{2}{*}{81.3}            & \multirow{2}{*}{6.14}    \\
                                   &                             &                                &                                &                               &                                  &                          \\
\multirow{2}{*}{Googlenet}         & \multirow{2}{*}{35.23\%}    & \multirow{2}{*}{37.70\%}       & \multirow{2}{*}{34.21\%}       & \multirow{2}{*}{35.24\%}      & \multirow{2}{*}{21.6}            & \multirow{2}{*}{5.97}    \\
                                   &                             &                                &                                &                               &                                  &                          \\
\multirow{2}{*}{Mobilenet-V2}      & \multirow{2}{*}{34.29\%}    & \multirow{2}{*}{38.21\%}       & \multirow{2}{*}{33.07\%}       & \multirow{2}{*}{34.29\%}      & \multirow{2}{*}{8.82}            & \multirow{2}{*}{5.13}    \\
                                   &                             &                                &                                &                               &                                  &                          \\
\multirow{2}{*}{T2T-ViT}           & \multirow{2}{*}{34.29\%}    & \multirow{2}{*}{38.17\%}       & \multirow{2}{*}{34.54\%}       & \multirow{2}{*}{34.28\%}      & \multirow{2}{*}{15.5}            & \multirow{2}{*}{4.44}    \\
                                   &                             &                                &                                &                               &                                  &                          \\
\multirow{2}{*}{Densenet169}       & \multirow{2}{*}{33.65\%}    & \multirow{2}{*}{36.55\%}       & \multirow{2}{*}{33.79\%}       & \multirow{2}{*}{33.65\%}      & \multirow{2}{*}{48.7}            & \multirow{2}{*}{11.13}   \\
                                   &                             &                                &                                &                               &                                  &                          \\
\multirow{2}{*}{InceptionResnetV1} & \multirow{2}{*}{33.64\%}    & \multirow{2}{*}{35.71\%}       & \multirow{2}{*}{32.90\%}       & \multirow{2}{*}{33.65\%}      & \multirow{2}{*}{30.9}            & \multirow{2}{*}{5.11}    \\
                                   &                             &                                &                                &                               &                                  &                          \\
\multirow{2}{*}{ResNet18}          & \multirow{2}{*}{33.33\%}    & \multirow{2}{*}{38.10\%}       & \multirow{2}{*}{32.36\%}       & \multirow{2}{*}{33.33\%}      & \multirow{2}{*}{42.7}            & \multirow{2}{*}{4.92}    \\
                                   &                             &                                &                                &                               &                                  &                          \\
\multirow{2}{*}{ResNet50}          & \multirow{2}{*}{33.33\%}    & \multirow{2}{*}{40.98\%}       & \multirow{2}{*}{33.44\%}       & \multirow{2}{*}{33.33\%}      & \multirow{2}{*}{90.1}            & \multirow{2}{*}{6.23}    \\
                                   &                             &                                &                                &                               &                                  &                          \\
\multirow{2}{*}{Densenet121}       & \multirow{2}{*}{33.01\%}    & \multirow{2}{*}{39.20\%}       & \multirow{2}{*}{33.79\%}       & \multirow{2}{*}{33.02\%}      & \multirow{2}{*}{27.1}            & \multirow{2}{*}{9.27}    \\
                                   &                             &                                &                                &                               &                                  &                          \\
\multirow{2}{*}{Deit}              & \multirow{2}{*}{32.39\%}    & \multirow{2}{*}{34.40\%}       & \multirow{2}{*}{32.74\%}       & \multirow{2}{*}{32.38\%}      & \multirow{2}{*}{21.1}            & \multirow{2}{*}{5.43}    \\
                                   &                             &                                &                                &                               &                                  &                          \\
\multirow{2}{*}{ViT}             & \multirow{2}{*}{31.75\%}    & \multirow{2}{*}{33.84\%}       & \multirow{2}{*}{31.47\%}       & \multirow{2}{*}{31.74\%}      & \multirow{2}{*}{31.2}            & \multirow{2}{*}{3.77}    \\
                                   &                             &                                &                                &                               &                                  &                          \\
\multirow{2}{*}{Inception-V3}      & \multirow{2}{*}{31.11\%}    & \multirow{2}{*}{34.84\%}       & \multirow{2}{*}{31.32\%}       & \multirow{2}{*}{31.11\%}      & \multirow{2}{*}{83.5}            & \multirow{2}{*}{7.49}    \\
                                   &                             &                                &                                &                               &                                  &                          \\
\multirow{2}{*}{ResNet101}         & \multirow{2}{*}{27.94\%}    & \multirow{2}{*}{34.59\%}       & \multirow{2}{*}{28.31\%}       & \multirow{2}{*}{27.94\%}      & \multirow{2}{*}{162}             & \multirow{2}{*}{8.83}    \\
                                   &                             &                                &                                &                               &                                  &                          \\
\multirow{2}{*}{VGG11}             & \multirow{2}{*}{27.61\%}    & \multirow{2}{*}{29.64\%}       & \multirow{2}{*}{26.00\%}       & \multirow{2}{*}{27.62\%}      & \multirow{2}{*}{491}             & \multirow{2}{*}{4.98}    \\
                                   &                             &                                &                                &                               &                                  &                          \\
\multirow{2}{*}{ShuffleNet-V2}     & \multirow{2}{*}{27.30\%}    & \multirow{2}{*}{25.02\%}       & \multirow{2}{*}{24.98\%}       & \multirow{2}{*}{27.30\%}      & \multirow{2}{*}{1.52}            & \multirow{2}{*}{5.42}    \\
                                   &                             &                                &                                &                               &                                  &                          \\
\multirow{2}{*}{BotNet}            & \multirow{2}{*}{25.40\%}    & \multirow{2}{*}{29.65\%}       & \multirow{2}{*}{26.04\%}       & \multirow{2}{*}{25.39\%}      & \multirow{2}{*}{72.2}            & \multirow{2}{*}{6.5}     \\
                                   &                             &                                &                                &                               &                                  &                          \\
\multirow{2}{*}{AlexNet}           & \multirow{2}{*}{24.44\%}    & \multirow{2}{*}{23.98\%}       & \multirow{2}{*}{22.65\%}       & \multirow{2}{*}{24.44\%}      & \multirow{2}{*}{217}             & \multirow{2}{*}{3.9}     \\
                                   &                             &                                &                                &                               &                                  &                          \\
\multirow{2}{*}{VGG13}             & \multirow{2}{*}{15.55\%}    & \multirow{2}{*}{15.18\%}       & \multirow{2}{*}{14.38\%}       & \multirow{2}{*}{15.55\%}      & \multirow{2}{*}{492}             & \multirow{2}{*}{5.28}    \\
                                   &                             &                                &                                &                               &                                  &                          \\
\multirow{2}{*}{VGG16}             & \multirow{2}{*}{8.26\%}     & \multirow{2}{*}{1.28\%}        & \multirow{2}{*}{1.93\%}        & \multirow{2}{*}{8.25\%}       & \multirow{2}{*}{512}             & \multirow{2}{*}{5.79}    \\
                                   &                             &                                &                                &                               &                                  &                          \\
\multirow{2}{*}{VGG19}             & \multirow{2}{*}{4.76\%}     & \multirow{2}{*}{0.23\%}        & \multirow{2}{*}{0.44\%}        & \multirow{2}{*}{4.76\%}       & \multirow{2}{*}{532}             & \multirow{2}{*}{6.42}    \\
                                   &                             &                                &                                &                               &                                  &                     \\    

\bottomrule
\end{tabular}}
\end{table*}

\subsection{Extended experiment}
\subsubsection{After Data Augmentation, the Classification Performance of Each Model on the Validation Set}

\begin{figure}
	\centering
		\includegraphics[scale=0.25]{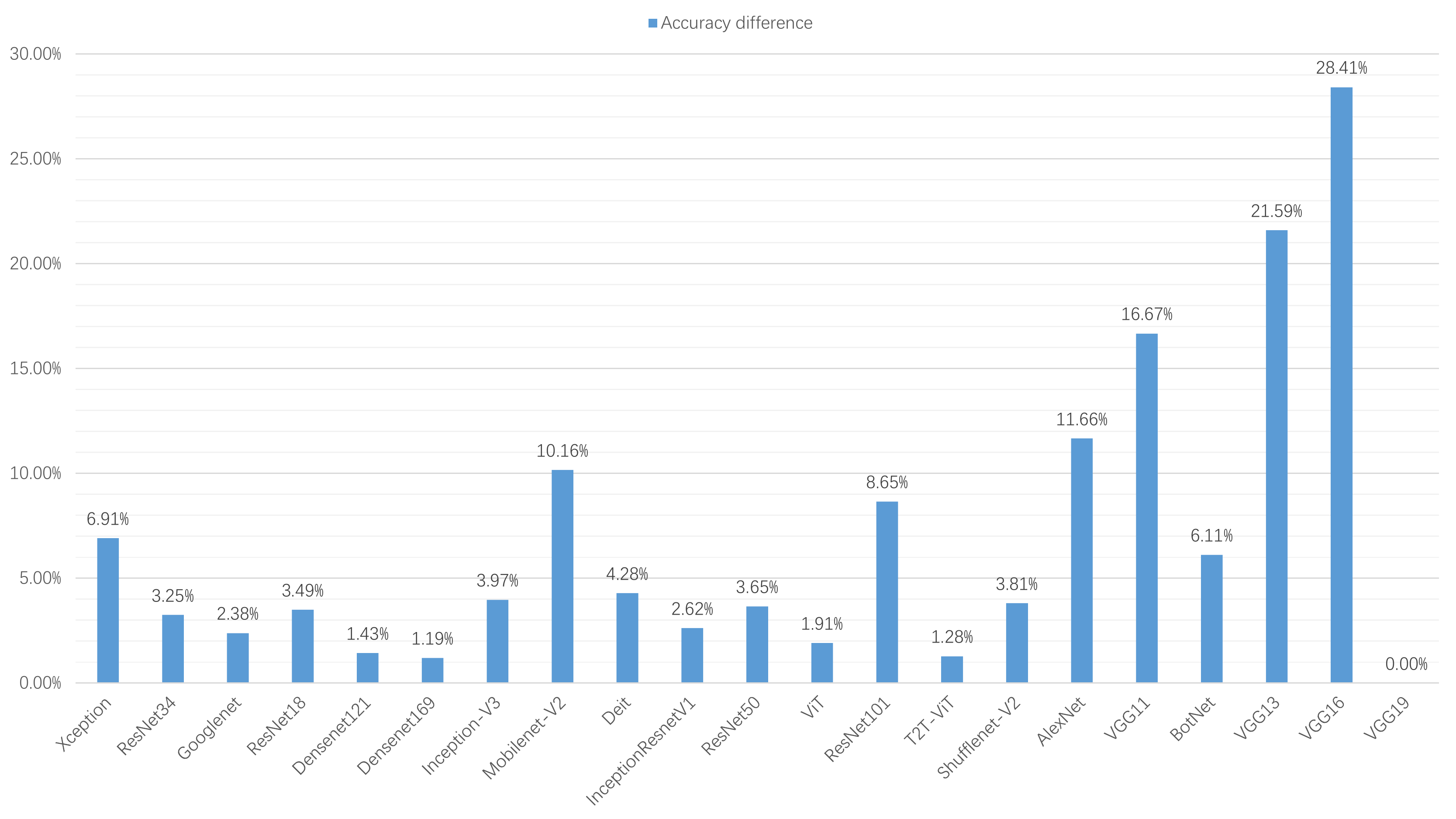}

	\caption{In the validation set of different deep learning model, the accuracy difference between data augmentation and before data augmentation.
}
	\label{FIG:6}
\end{figure}

In this section, we augment the dataset, and the performance indicators of the models are calculated and exhibited in Table. \ref{tbl8}, including precision, recall, F1-score, and accuracy. In addition, we compare the accuracy changes before and after data augmentation, as shown in Figure. \ref{FIG:6}.

After data augmentation, the time required for model training also increases significantly. The training time of the ViT models is the least, which is 902.27s. Although the training set is augmented to six times, the training time of the ViT models is increased by 187.27s compared with the 715s. The classification accuracy of the Xception network ranks first at 52.62\%. The T2T-ViT network has the lowest classification rate of 35.56\%.

After data augmentation, the classification performance of each model is improved. Figure.~\ref{FIG:6} shows the changes in the accuracy of each model after data augmentation. The validation set accuracy of the VGG16 network is increased the most, at 28.41\%. This is because the VGG16 network can converge on the augmentation dataset. In addition, the validation set accuracy of VGG13 and VGG11 are improved significantly, increasing by 21.59\% and 16.67\%, respectively. The accuracy of the VGG11 validation set rose from 17th to 3th. The accuracy of the VGG13 validation set rose from 19th to 11th. After data augmentation, the validation set accuracy of T2T-ViT, Densenet169 and ViT are not improved significantly, increasing by 1.28\%, 1.19\%, and 1.91\%.

From a specific series of models, the performance of VGG series models is improved significantly after data augmentation. The performance improvement of the Densenet series models is not apparent. The accuracy of the Densenet121 and the Densenet169 validation sets are increased by 1.43\% and 1.19\%, respectively. Meanwhile, the performance improvement of the VT series models is not apparent. The classification accuracy of the T2T-ViT validation set is increased by 1.28\%, ViT is increased by 1.91\%, and Diet is increased by 4.28\%. In the ResNet series models, ResNet18, ResNet34 and ResNet50 are increased by 3.49\%, 3.25\% and 3.65\%, and the improvement is not obvious. However, the classification accuracy of the ResNet101 validation set is increased by 8.65\%, which is obvious.

\begin{table*}
\caption{Comparison of classification results of different deep learning models in the validation set. P denotes Precision, and R represents Recall. The training set is augmented. (Sort in descending order of classification accuracy.)}\label{tbl8}
\renewcommand\arraystretch{1.2} 
\centering
\setlength{\tabcolsep}{3.2mm}{
\begin{tabular}{@{}ccccccccccc@{} }
\toprule

Model             & Avg. R(\%) & Avg. P(\%) & Avg. F1\_score(\%) & Accuracy(\%) & Params Size (MB) & Time (s) \\   \midrule
Xception          & 52.62\%       & 52.05\%    & 50.63\%       & 52.62\%      & 79.80            & 2636.08  \\
Mobilenet-V2      & 49.67\%       & 51.91\%    & 48.82\%       & 49.68\%      & 8.82             & 1237.49  \\
VGG11             & 48.10\%       & 52.40\%    & 48.44\%       & 48.10\%      & 491.00           & 1745.73  \\
ResNet34          & 46.10\%       & 47.85\%    & 44.68\%       & 46.11\%      & 81.30            & 1335.87  \\
ResNet18          & 44.44\%       & 51.87\%    & 43.03\%       & 44.44\%      & 42.70            & 1090.39  \\
Googlenet         & 44.29\%       & 47.16\%    & 43.50\%       & 44.29\%      & 21.60            & 1257.33  \\
Inception-V3      & 43.97\%       & 50.78\%    & 43.41\%       & 43.97\%      & 83.50            & 2004.08  \\
AlexNet           & 43.58\%       & 45.02\%    & 43.05\%       & 43.57\%      & 217.00           & 951.27   \\
ResNet101         & 43.41\%       & 46.08\%    & 43.33\%       & 43.41\%      & 162.00           & 2786.95  \\
Deit              & 43.34\%       & 46.62\%    & 43.29\%       & 43.33\%      & 21.10            & 1306.99  \\
VGG13             & 42.54\%       & 41.38\%    & 41.21\%       & 42.54\%      & 492.00           & 2307.04  \\
Densenet121       & 42.38\%       & 46.91\%    & 42.39\%       & 42.38\%      & 27.10            & 2169.11  \\
ResNet50          & 42.22\%       & 47.76\%    & 42.10\%       & 42.22\%      & 90.10            & 1968.28  \\
Densenet169       & 42.14\%       & 48.04\%    & 42.79\%       & 42.14\%      & 48.70            & 2526.61  \\
InceptionResnetV1 & 41.66\%       & 47.83\%    & 41.68\%       & 41.67\%      & 30.90            & 1451.76  \\
ViT               & 39.05\%       & 43.50\%    & 38.52\%       & 39.05\%      & 31.20            & 902.27   \\
ShuffleNet-V2       & 37.62\%       & 39.37\%    & 36.84\%       & 37.62\%      & 1.52             & 965.81   \\
VGG16             & 37.47\%       & 38.21\%    & 36.80\%       & 37.46\%      & 512.00           & 2589.15  \\
BotNet            & 36.59\%       & 36.38\%    & 35.59\%       & 36.59\%      & 72.20            & 2000.17  \\
T2T-ViT           & 35.56\%       & 38.43\%    & 36.19\%       & 35.56\%      & 15.50            & 1385.62  \\
VGG19             & 4.76\%        & 0.23\%     & 0.44\%        & 4.76\%       & 532.00           & 1022.57 \\
\bottomrule
\end{tabular}}
\end{table*}

\subsubsection{After Data Augmentation, the Classification Performance of Each Model on the Test Set}

After data augmentation, the performance of each model on the test set is shown in Table. \ref{tbl9}. In Table. \ref{tbl9}, the Xception network has the highest accuracy of 45.71\%. Meanwhile, the Xception network has an excellent recall index of 50.43\%. Excluding the non-convergent VGG19, the VGG16 model has the worst performance, with an accuracy of 24.76\%. The ViT model consumes the least time, which is 3.72s. The Densenet169 model consumes the most time, which is 11.04s.

\begin{figure}
	\centering
		\includegraphics[scale=0.25]{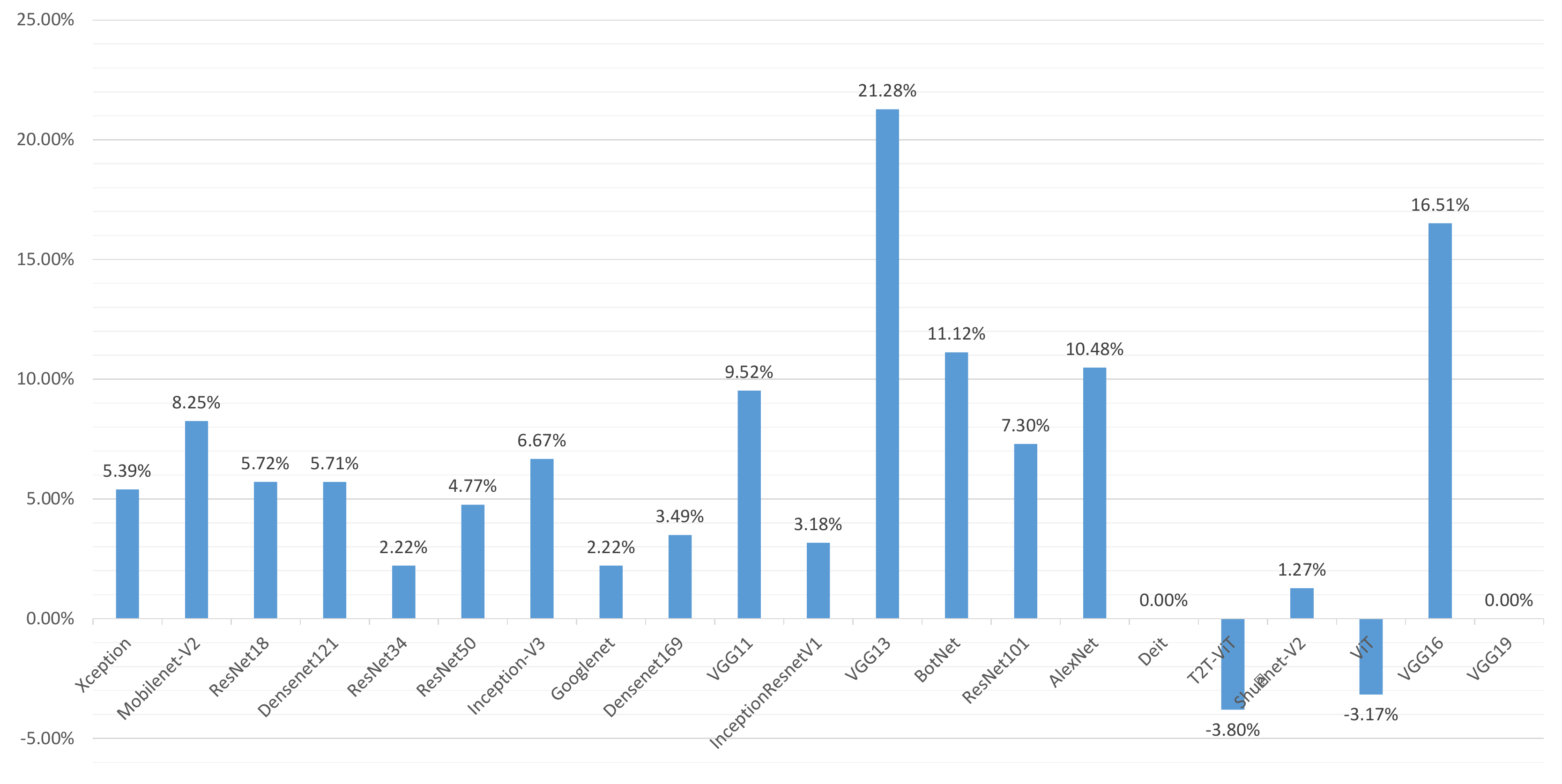}

	\caption{In the test set of different deep learning model, the accuracy difference between data augmentation and before data augmentation.
}
	\label{FIG:7}
\end{figure}

Figure.~\ref{FIG:7} shows the change of accuracy on the test set before and after the augmentation. In Figure.~\ref{FIG:7}, we can see that the accuracy of each deep learning model on the test set is generally increased. Among them, the accuracy of the VGG series models is improved the most. VGG11 is increased by 9.25\%, VGG13 is increased by 21.28\%, and VGG16 is increased by 16.51\%. However, the accuracy of the VT series models test set is not significantly improved. The accuracy of some model test sets even drops. After data augmentation, the accuracy of the Diet network validation set is not changed. The accuracy of the T2T-ViT network is dropped by 3.80\%. The accuracy of the ViT model is dropped by 3.17\%. However, the accuracy of BotNet, a mixed model of CNN and VT, is improved significantly, reaching 11.12\%.

\begin{table*}
\caption{Comparison of classification results of different deep learning models in the test set. P denotes Precision, and R represents Recall. The training set is augmented. (Sort in descending order of classification accuracy.)}\label{tbl9}
\renewcommand\arraystretch{1.1} 
\centering
\setlength{\tabcolsep}{3.2mm}{
\begin{tabular}{@{}ccccccccccc@{} }
\toprule

Model             & Avg. R(\%) & Avg. P(\%) & Avg. F1\_score(\%) & Accuracy(\%) & Params Size (MB) & Time (s) \\   \midrule
Xception          & 45.71\%       & 50.43\%    & 46.15\%       & 45.71\%      & 79.8             & 5.49     \\
Mobilenet-V2      & 42.54\%       & 47.56\%    & 43.07\%       & 42.54\%      & 8.22             & 5.04     \\
ResNet18          & 39.05\%       & 44.82\%    & 39.22\%       & 39.05\%      & 42.7             & 4.90     \\
Densenet121       & 38.73\%       & 40.28\%    & 38.20\%       & 38.73\%      & 27.1             & 8.98     \\
ResNet34          & 38.73\%       & 42.25\%    & 37.84\%       & 38.73\%      & 81.3             & 6.07     \\
ResNet50          & 38.10\%       & 41.56\%    & 36.97\%       & 38.10\%      & 90.1             & 6.20     \\
Inception-V3      & 37.78\%       & 44.32\%    & 38.00\%       & 37.78\%      & 83.5             & 7.47     \\
Googlenet         & 37.46\%       & 43.55\%    & 37.92\%       & 37.46\%      & 21.6             & 6.03     \\
Densenet169       & 37.14\%       & 41.51\%    & 37.37\%       & 37.14\%      & 48.7             & 11.04    \\
VGG11             & 37.14\%       & 38.81\%    & 36.70\%       & 37.14\%      & 491              & 4.96     \\
InceptionResnetV1 & 36.82\%       & 41.47\%    & 36.75\%       & 36.83\%      & 30.9             & 5.11     \\
VGG13             & 36.82\%       & 38.46\%    & 36.25\%       & 36.83\%      & 492              & 5.28     \\
BotNet            & 36.50\%       & 39.12\%    & 36.35\%       & 36.51\%      & 72.2             & 6.44     \\
ResNet101         & 35.23\%       & 38.01\%    & 35.44\%       & 35.24\%      & 162              & 8.85     \\
AlexNet           & 34.92\%       & 39.10\%    & 34.97\%       & 34.92\%      & 217              & 5.25     \\
Deit              & 32.39\%       & 34.40\%    & 32.74\%       & 32.38\%      & 21.1             & 4.41     \\
T2T-ViT           & 30.48\%       & 35.88\%    & 30.85\%       & 30.48\%      & 15.50          & 5.41     \\
ShuffleNet-V2        & 28.57\%       & 35.64\%    & 29.41\%       & 28.57\%      & 1.52             & 5.42     \\
ViT               & 28.58\%       & 29.63\%    & 27.86\%       & 28.57\%      & 31.2             & 3.72     \\
VGG16             & 24.77\%       & 25.53\%    & 24.11\%       & 24.76\%      & 512              & 5.79     \\
VGG19             & 4.76\%        & 0.23\%     & 0.44\%        & 4.76\%       & 532              & 6.36    \\
\bottomrule
\end{tabular}}
\end{table*}

\subsubsection{In Imbalanced Training, after Data Augmentation, the Classification Performance of Each Model on the Validation Set}

\begin{table*}
\caption{AP and MAP of different deep learning models in imbalanced  training. (In [\%].)}\label{tbl10}
\renewcommand\arraystretch{1.2} 

\begin{tabular}{@{}cccccccccccc@{} }
\toprule

model/sample & 1       & 2       & 3       & 4       & 5       & 6       & 7       & 8       & 9       & 10      & 11      \\  \midrule
ViT                         & 30.77\% & 44.99\% & 18.43\% & 48.51\% & 74.47\% & 76.17\% & 50.98\% & 15.32\% & 31.12\% & 60.74\% & 54.02\% \\
Xception                    & 37.66\% & 51.16\% & 29.72\% & 68.32\% & 73.66\% & 67.96\% & 79.19\% & 65.41\% & 55.84\% & 82.97\% & 55.91\% \\
VGG16                       & 48.38\% & 41.43\% & 9.63\%  & 51.05\% & 52.61\% & 42.23\% & 76.92\% & 5.97\%  & 27.57\% & 76.12\% & 34.77\% \\
ResNet50                    & 30.58\% & 45.96\% & 14.24\% & 68.19\% & 66.15\% & 43.10\% & 71.24\% & 46.51\% & 31.87\% & 62.19\% & 36.79\% \\
Inception-V3                & 37.75\% & 36.79\% & 33.41\% & 56.37\% & 55.77\% & 43.51\% & 59.52\% & 41.18\% & 38.40\% & 75.03\% & 69.26\% \\  \midrule
model/sample & 12      & 13      & 14      & 15      & 16      & 17      & 18      & 19      & 20      & 21      & mAP     \\   \midrule
ViT                         & 15.24\% & 17.84\% & 25.46\% & 6.74\%  & 13.95\% & 48.61\% & 7.26\%  & 60.33\% & 23.07\% & 9.53\%  & 34.93\% \\
Xception                    & 54.16\% & 52.28\% & 65.06\% & 46.36\% & 30.61\% & 60.41\% & 31.21\% & 61.14\% & 45.50\% & 74.36\% & 56.61\% \\
VGG16                       & 24.06\% & 16.22\% & 63.90\% & 5.80\%  & 10.49\% & 33.87\% & 24.77\% & 44.00\% & 33.14\% & 5.47\%  & 34.69\% \\
ResNet50                    & 15.59\% & 42.12\% & 68.57\% & 24.94\% & 17.49\% & 47.52\% & 6.64\%  & 49.04\% & 16.73\% & 56.10\% & 41.03\% \\
Inception-V3                & 15.09\% & 49.09\% & 64.11\% & 37.91\% & 15.00\% & 43.98\% & 15.84\% & 54.40\% & 10.78\% & 60.38\% & 43.50\% \\

\bottomrule
\end{tabular}
\end{table*}

In this section, we re-split and combine the data. Take each of the 21 types of \emph{environmental microorganisms} (EMs) as positive samples in turn and the remaining 20 types of microorganisms as negative samples. In this way, we repeat this process 21 times in our paper. The specific splitting method is shown in section \ref{sec:Data settings}. The deep learning model can calculate an AP after training each piece of data. Table \ref{tbl10} shows the AP and mAP of each model validation set. We select the classical VGG16, ResNet50 and Inception-V3 networks for experiments. Furthermore, a relatively novel ViT model is also selected. In addition, the Xception network, which has always performed well above, is selected for experiments. Since the VGG16 network cannot converge at a learning rate of 0.0001, this part of the experiment adjusts the learning rate of the VGG16 network to 0.00001.

\begin{figure*}
	\centering
		\includegraphics[scale=0.44]{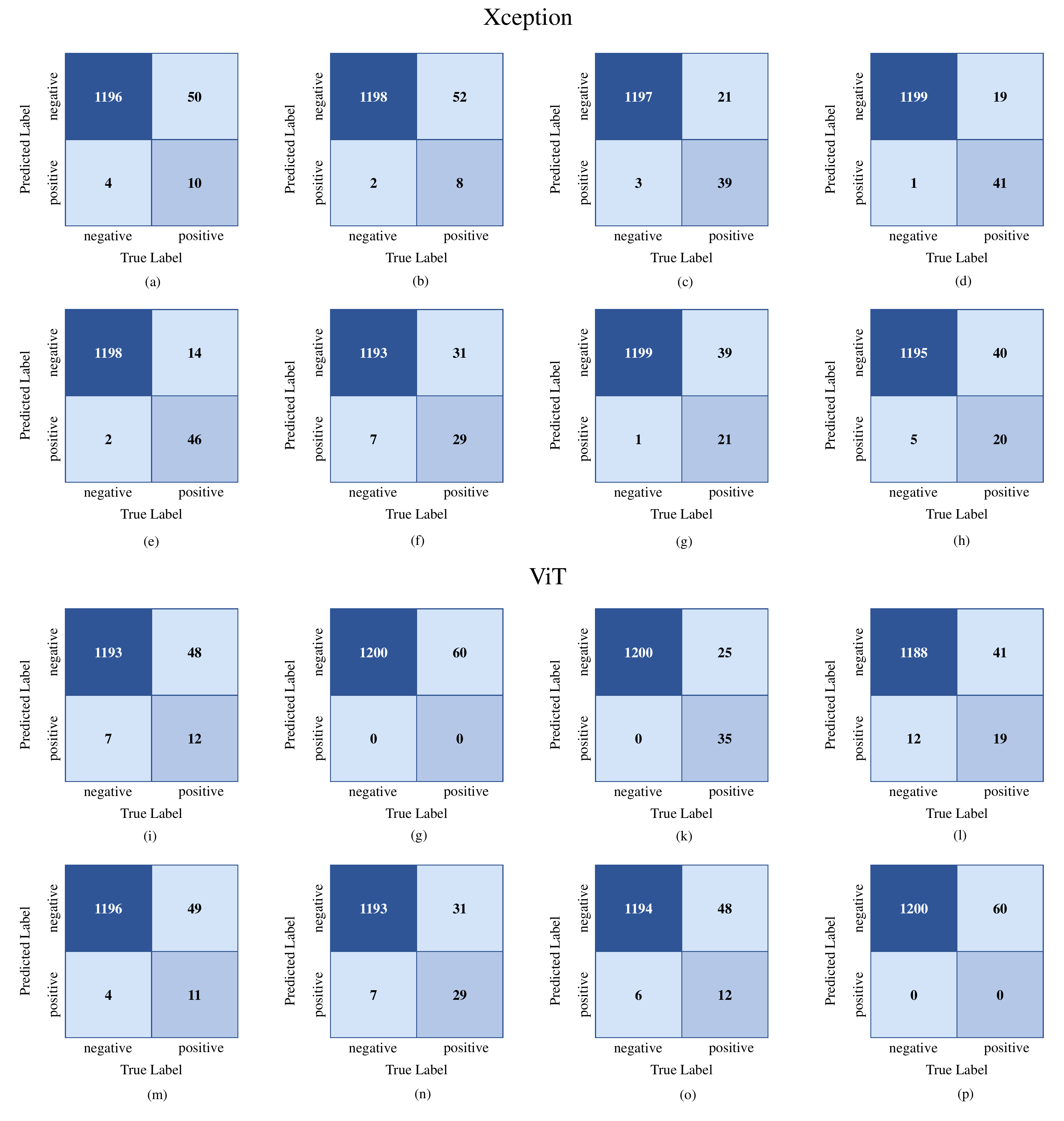}

	\caption{The confusion matrices a, b, c, d, e, f, g, and h are drawn based on the Xception network validation set results. Likewise, the confusion matrices i, j, k, l, m, n, o and p are drawn based on the ViT network validation set results. a, b, c, d, e, f, g and h are generated from datasets 1, 3, 5, 7, 10, 11, 13, 15. i, j, k, l, m, n, o, and p are generated from datasets 1, 3, 6, 7, 9, 11, 13, and 15, respectively. (Dataset segmentation is shown in~\ref{sec:Data settings} Experiment B.)
}
	\label{FIG:8}
\end{figure*}

It can be seen in Table \ref{tbl10} that the mAP of the Xception network is the highest, which is 56.61\%. The Xception network has the highest AP on the 10th data, and the AP is 82.97\%. The Xception network has the worst AP on the 3rd data, with an AP of 29.72\%. As shown in Figure \ref{FIG:8}, the confusion matrix (d) is drawn by the 10th data. In (d), 46 of the 60 positive samples are classified correctly, and 14 are mistakenly classified as negative samples. In the confusion matrix drawn by the third data, 8 of the 60 positive samples are classified correctly, and 52 are incorrectly classified as negative samples.

The mAp of the VGG16 network is the lowest at 34.69\%. The VGG16 network performs best on the 10th data AP, with an AP of 76.12\%. The VGG16 network performs the worst on the 21st data AP, with an AP of 5.47\%. Despite adjusting the learning rate, the VGG16 network still fails to converge on the 3rd, 8th, 13th, 15th, and 21st data.

The mAp of the ViT network and the VGG16 network are relatively close. The ViT network performs best on the 6th data AP, with an AP of 76.17\%. Among the 60 positive samples, 35 are classified correctly, and 25 are classified as negative samples. The ViT network performs the worst on the 15th data AP, with an AP of 6.74\%. Among the 60 positive samples, 0 are classified correctly and 60 are classified as negative samples.

In addition, Resnet50 performs the best on 7 data AP and the worst on the 18th data AP. The Inception-V3 network performs best on 10 data AP and the worst on the 16th data AP.

\subsubsection{Mis-classification Analysis}

\begin{figure*}
	\centering
		\includegraphics[scale=0.35]{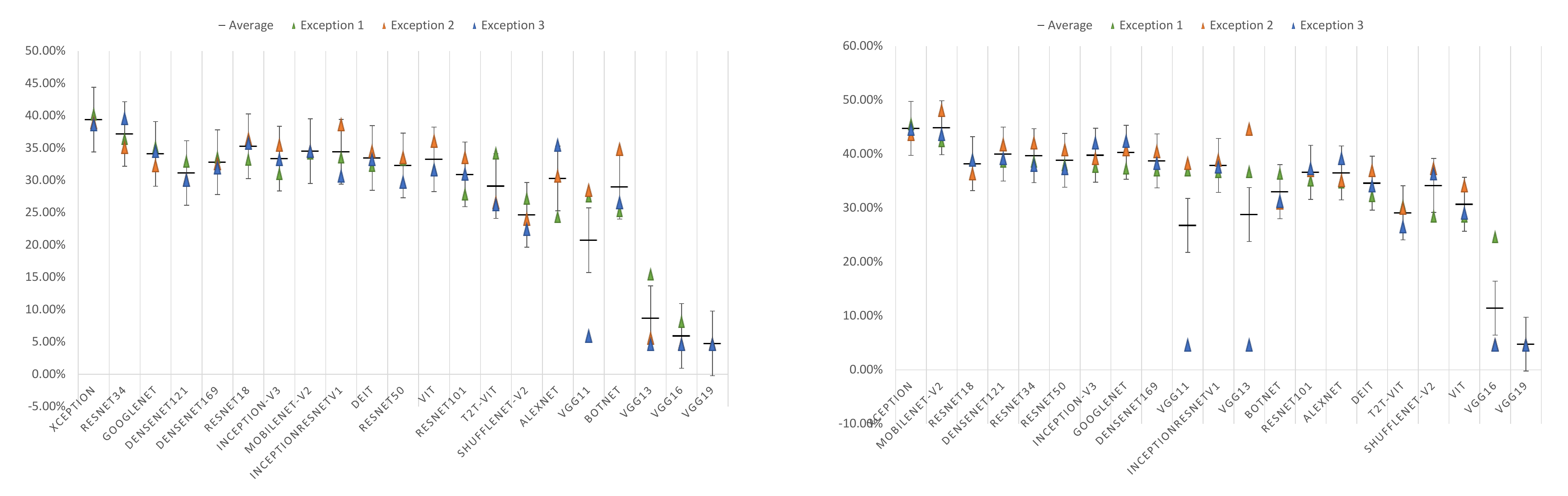}

	\caption{Error bar of accuracy on test set. The left figure shows the test set error bar before data expansion. The figure on the right shows the error bar of the test set after data expansion.}

	\label{ErrorBar}
\end{figure*}

In the extended experiments, we randomly divide the EMDS-6 dataset three times and train the data for each division. The results and accuracy errors of the three experiments are shown in Table.~\ref{Average} and Figure.~\ref{ErrorBar}.

\begin{figure}
	\centering
		\includegraphics[scale=0.25]{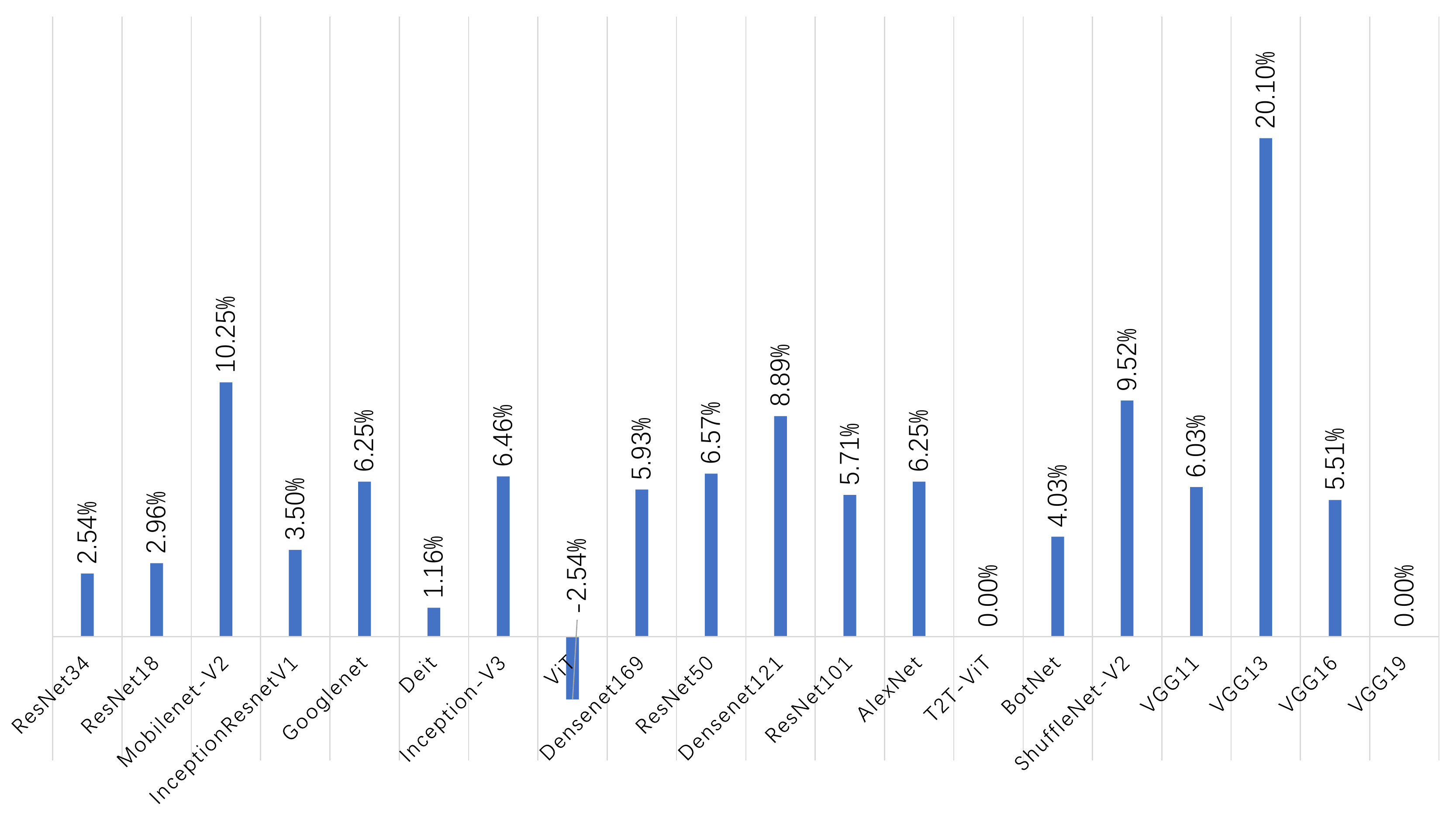}

	\caption{Error bar of accuracy on test set. The left figure shows the test set error bar before data expansion. The figure on the right shows the error bar of the test set after data expansion.}

	\label{difference}
\end{figure}

In Table.~\ref{Average}, under the original dataset, Xception has the best classification performance on 21 deep learning models. After data augmentation, Xception still has the highest classification performance. In Table.~\ref{Average}, the performance of the VGG series network has major changes compared to Table.~\ref{tbl9}. In Figure.~\ref{difference}, we can clearly understand that VGG11, VGG13, VGG16, and VGG19 failed to converge at least once in the three experiments. This phenomenon causes the VGG series models to fall behind in average performance. Except for the VGG series models, the performance of other models tends to be stable on the whole, and the errors are kept within ± $5\%$ of the average of the three experiments. Xception and Densenet169 networks show good robustness in the classification results before and after data augmentation. However, the classification performance of the AlexNet network fluctuates greatly in the three experiments, and the robustness is poor.

In Figure .~\ref{difference}, after data augmentation, the performance of VGG13 improves the most, but this is mainly caused by the failure of some experiments on the original dataset to converge. In addition to the VGG13 network, the Mobilenet-V2, ShuffleNet-V2, and Densenet121 models improve the most, with accuracy rates increase by 10.25\%, 9.52\%, and 8.89\%. In addition, the performance improvement of ResNet34, ResNet18, and InceptionResnetV1 models is relatively small, and the accuracy are increase by 2.54\%, 2.96\%, and 3.5\%. Generally speaking, after data augmentation, the CNN series models have a very obvious improvement in the precision, recall, F1-Score, and accuracy of the test set. However, the opposite situation appeared in the VTs after data augmentation. Taking the Accuracy index as an example, the accuracy of the ViT model in the test set has dropped by -2.5\%, the Accuracy of the T2T-ViT model is equal to that before the augmentation, and the Accuracy of the Deit model has only increased by 1.16\%.

In general, augmenting the dataset through geometric transformation can effectively improve the classification performance of the CNN series models. Nevertheless, for the VTs, the method of geometric transformation to augment the dataset is difficult to improve the classification performance of the VTs and even leads to a decrease in model performance.



\begin{table*}[]
\renewcommand\arraystretch{1} 
\caption{Comparison of different deep learning models on test sets. (In [\%].)}\label{Average}
\setlength{\tabcolsep}{2.9mm}{
\begin{tabular}{llllllllll}
\bottomrule
\textbf{Mode1}     & \multicolumn{2}{l}{\textbf{Original Data }} & & & & \multicolumn{2}{l}{\textbf{Augmented Data }}&     \\
\cmidrule(lr){2-5}\cmidrule(lr){7-10}
& Recall &  Precision  & F1-Score & Accuracy & &Recall & Precision  & F1-Score & Accuracy\\
\bottomrule
Xception          & 39.37\% & 44.25\% & 39.07\% & 39.37\% &  & 44.76\% & 47.97\% & 44.53\% & 44.76\% \\
ResNet34          & 37.14\% & 41.96\% & 36.93\% & 37.14\% &  & 39.68\% & 43.15\% & 39.54\% & 39.68\% \\
ResNet18          & 35.24\% & 40.53\% & 34.33\% & 35.24\% &  & 38.20\% & 42.64\% & 38.38\% & 38.20\% \\
Mobilenet-V2      & 34.50\% & 37.24\% & 33.86\% & 34.50\% &  & 44.75\% & 48.31\% & 44.82\% & 44.75\% \\
InceptionResnetV1 & 34.39\% & 36.46\% & 33.92\% & 34.39\% &  & 37.88\% & 41.09\% & 37.53\% & 37.89\% \\
Googlenet         & 34.07\% & 36.89\% & 33.48\% & 34.07\% &  & 40.32\% & 44.59\% & 40.37\% & 40.32\% \\
Deit              & 32.27\% & 34.08\% & 31.92\% & 33.44\% &  & 34.60\% & 37.01\% & 34.76\% & 34.60\% \\
Inception-V3      & 33.33\% & 33.78\% & 32.26\% & 33.33\% &  & 39.79\% & 43.17\% & 39.69\% & 39.79\% \\
ViT               & 33.24\% & 34.92\% & 32.63\% & 33.23\% &  & 30.69\% & 32.49\% & 30.08\% & 30.69\% \\
Densenet169       & 32.80\% & 35.38\% & 32.49\% & 32.80\% &  & 38.73\% & 43.52\% & 38.79\% & 38.73\% \\
ResNet50          & 32.28\% & 36.41\% & 31.79\% & 32.27\% &  & 38.84\% & 41.69\% & 38.37\% & 38.84\% \\
Densenet121       & 31.11\% & 35.66\% & 31.25\% & 31.11\% &  & 40.00\% & 43.02\% & 39.75\% & 40.00\% \\
ResNet101         & 30.90\% & 35.29\% & 30.97\% & 30.90\% &  & 36.61\% & 38.34\% & 36.01\% & 36.61\% \\
AlexNet           & 30.26\% & 31.08\% & 28.70\% & 30.26\% &  & 36.51\% & 39.62\% & 36.41\% & 36.51\% \\
T2T-ViT           & 29.10\% & 32.84\% & 29.17\% & 29.10\% &  & 29.10\% & 32.19\% & 29.13\% & 29.10\% \\
BotNet            & 29.00\% & 31.11\% & 28.46\% & 28.99\% &  & 33.02\% & 34.29\% & 32.45\% & 33.02\% \\
ShuffleNet-V2     & 24.66\% & 23.71\% & 22.86\% & 24.66\% &  & 34.18\% & 37.09\% & 34.19\% & 34.18\% \\
VGG11             & 20.74\% & 19.99\% & 18.31\% & 20.74\% &  & 26.77\% & 26.98\% & 25.39\% & 26.77\% \\
VGG13             & 8.68\%  & 5.66\%  & 5.47\%  & 8.68\%  &  & 28.78\% & 29.52\% & 27.12\% & 28.78\% \\
VGG16             & 5.93\%  & 0.58\%  & 0.94\%  & 5.92\%  &  & 11.43\% & 8.66\%  & 8.33\%  & 11.43\% \\
VGG19             & 4.76\%  & 0.23\%  & 0.44\%  & 4.76\%  &  & 4.76\%  & 0.23\%  & 0.44\%  & 4.76\%\\
\bottomrule
\end{tabular}}
\end{table*}

\subsubsection{Comparison of Experimental Results after Adjusting Model Parameters}

\begin{table*}[]
\renewcommand\arraystretch{1.17} 
\caption{Comparison of training time consumption and test set accuracy of different models. The left side of the table shows the training time consumption, while the right side of the table shows the accuracy of the test set. learning rate (LR), Batch Size (BS).}\label{time_acc}
\setlength{\tabcolsep}{4.3mm}{
\begin{tabular}{@{}ccccccccccc@{} }
\hline
\multirow{2}{*}{\textbf{LR}} & \multicolumn{4}{c}{\textbf{ViT (Times)}}      & \textbf{} & \multicolumn{4}{c}{\textbf{ViT (Accuracy)}}      \\ \cline{2-5} \cline{7-10} 
                             & BS 4      & BS 8      & BS 16     & BS 32     &           & BS 4        & BS 8       & BS 16     & BS 32     \\ \cline{1-5} \cline{7-10} 
$2 \times 10^{-5}$                    & 530.70    & 793.56    & 760.76    & 760.99    &           & 28.25\%     & 21.59\%    & 16.83\%   & 14.92\%   \\
$2 \times 10^{-4}$                        & 530.32    & 793.12    & 761.17    & 762.88    &           & 30.16\%     & 27.94\%    & 31.11\%   & 30.16\%   \\
$2 \times 10^{-3}$                    & 530.30    & 792.43    & 760.87    & 761.88    &           & 11.43\%     & 15.87\%    & 20.63\%   & 17.14\%   \\
$2 \times 10^{-2}$                       & 535.30    & 794.01    & 760.67    & 760.99    &           & 4.76\%      & 4.76\%     & 3.17\%    & 7.62\%    \\ \cline{1-5} \cline{7-10} 
\multirow{2}{*}{\textbf{LR}} & \multicolumn{4}{c}{\textbf{Xception (Times)}} & \textbf{} & \multicolumn{4}{c}{\textbf{Xception (Accuracy)}} \\ \cline{2-5} \cline{7-10} 
                             & BS 4      & BS 8      & BS 16     & BS 32     &           & BS 4        & BS 8       & BS 16     & BS 32     \\ \cline{1-5} \cline{7-10} 
$2 \times 10^{-5}$                        & 840.31    & 1106.94   & 1119.99   & 1074.65   &           & 38.10\%     & 37.46\%    & 37.14\%   & 37.78\%   \\
$2 \times 10^{-4}$                    & 834.88    & 1107.95   & 1081.05   & 1088.86   &           & 51.43\%     & 50.48\%    & 41.90\%   & 38.73\%   \\
$2 \times 10^{-3}$                   & 837.11    & 1113.56   & 1042.63   & 1042.75   &           & 23.49\%     & 34.29\%    & 28.25\%   & 30.48\%   \\
$2 \times 10^{-2}$                         & 808.83    & 1086.24   & 1037.36   & 1073.62   &           & 14.29\%     & 16.83\%    & 20.00\%   & 17.46\%   \\ \cline{1-5} \cline{7-10} 
\multirow{2}{*}{\textbf{LR}} & \multicolumn{4}{c}{\textbf{BotNet (Times)}}   & \textbf{} & \multicolumn{4}{c}{\textbf{BotNet (Accuracy)}}   \\ \cline{2-5} \cline{7-10} 
                             & BS 4      & BS 8      & BS 16     & BS 32     &           & BS 4        & BS 8       & BS 16     & BS 32     \\ \cline{1-5} \cline{7-10} 
$2 \times 10^{-5}$                      & 806.80    & 1006.82   & 977.10    & 950.71    &           & 16.51\%     & 17.14\%    & 20.32\%   & 21.27\%   \\
$2 \times 10^{-4}$                  & 778.31    & 990.82    & 1022.45   & 1011.02   &           & 12.70\%     & 24.76\%    & 26.67\%   & 27.94\%   \\
$2 \times 10^{-3}$                        & 772.63    & 984.33    & 967.09    & 937.65    &           & 9.21\%      & 7.94\%     & 14.29\%   & 10.79\%   \\
$2 \times 10^{-2}$                        & 774.33    & 985.43    & 968.46    & 936.39    &           & 7.94\%      & 10.79\%    & 7.62\%    & 16.83\%   \\ \hline
\end{tabular}}
\end{table*}

In this section, our extended experiments select representative models, namely the CNN-based Xception, the Transformer-based ViT, and the BotNet hybrid model based on CNN and VT. This section of the experiment trains 100 epochs. The purpose of the study is to observe the effect of changing two hyperparameters, learning rate (LR) and batch size (BS), on the experimental results. The experimental results are shown in Table \ref{time_acc}.

Under the same BS and different Learning Rates (LRs) conditions, the maximum fluctuation of ViT training time is only 4.6s, the maximum fluctuation of BotNet training time is 74.6s, and the maximum fluctuation of Xceotion training time is 80.6s. Experiments indicate that adjusting LRs has little effect on the time required for training. However, the change of LRs greatly influences the accuracy of experimental results. Taking the ViT as an example, the accuracy of the model is 16.83\% under the conditions of BS=16 and LR=$2 \times 10^{-5}$. Under the condition of LR=$2 \times 10^{-4}$, the highest accuracy of the ViT can reach 31.11\%. In addition, the accuracy of the model is only 3.17\% under the condition of LR=$2 \times 10^{-2}$. Experiments indicate that the performance of the model decreases when using an oversized LR (LR=$2 \times 10^{-2}$) and an extremely small LR (LR=$2 \times 10^{-5}$). An oversized LR may cause the network to fail to converge, which means the model lingered near the optimal value and could not reach the optimal solution. This leads to performance degradation. The following two reasons explain the performance degradation when applying extremely small LRs. On the one hand, an extremely small LR makes the network hard to converge fastly. The related experiments show that the model is difficult to reach the optimal value within 100 epochs with an extremely small LR ($2 \times 10^{-5}$). On the other hand, an extremely small LR may cause the network to fall into an optimal local solution, which leads to performance degradation.

In addition to the LR, the change of BS also dramatically affects the performance of the model. Different models show different patterns at different BS values. For example, the accuracy of the ViT model decreases rapidly with increasing BS at LR=$2 \times 10^{-5}$. The accuracy of the BotNet increases sharply with increasing BS at LR=$2 \times 10^{-5}$. However, the relevant experiments show that BS does not seriously affect the performance of the model under large datasets~\cite{radiuk_2017_impact}. Nevertheless, with small datasets, only a slight change in the BS value can dramatically change the performance of the model.

Compared to a large dataset, adjusting the BS and learning rate on a small dataset can significantly change model performance. Therefore, finding the optimal parameters to improve the model performance on small datasets is necessary.

\subsection{Discussion}

This experiment studies the classification performance of 21 deep learning models on  small EM dataset (EMDS-6). The comparison results are obtained according to the evaluation indicators, as shown in Tables \ref{tbl6}, \ref{tbl7}, \ref{tbl8} and \ref{tbl9}. Meanwhile, some models are selected for imbalanced experiments to investigate the performance of the models further. The results are shown in Table \ref{tbl10}. In order to increase the reliability of the conclusions, this paper repeats the main experiment three times. The average value is shown in Table~\ref{Average}, and the errors of the three experiments are shown in Figure~\ref{ErrorBar}. In addition, this paper explores the impact of hyperparameters on small dataset classification, and the results are shown in Table ~\ref{time_acc}.

The performance of the VGG network gradually decreases as the number of network layers increases. Especially the VGG16 and VGG19 networks cannot converge on the EMDS-6 dataset. This may be because the dataset is too small, and the gradient disappears in the process of a continuous deepening of the network layer, which affects the convergence.

The training time of the ViT network on the EMDS-6 database is very short, but it does not make a significant difference with other models. After the data augmentation of EMDS-6, the ViT network has apparent advantages in the time of training the model, and the time consumption is much less than other models. We can speculate that the ViT model may further expand its advantage when trained on more training data.

In the experiments where the model parameters are tuned, slight changes in both the LR and BS parameters lead to drastic changes in model performance. This does not happen if the experiment is based on a large-scale dataset. However, in small datasets, each class of EMs only accounts for a portion of the image, and most of the others are noise. Moreover, some models that include batch normalization normalize the environmental noise at different BS leading to fluctuations in classification accuracy.

After data augmentation, the accuracy of CNN series models improves significantly. However, the increase of VT series model accuracy is slight, and some of them even decrease. The results are shown in Figure~\ref{FIG:7}. To further prove the above experimental results, this paper re-divides the dataset and conducts three experiments, and the results are shown in Figure~\ref{difference}. Experiments once again prove that the geometric deformation augmented data method is difficult to improve the performance of the VT series models. This may be because our data augmentation method only makes geometric changes to the data. The geometric transformation is only changed the spatial position of the feature. However, the VT series models use attention to capture the global context information, and it pays more attention to global information. Operations such as rotation and mirroring have little effect on global information, and it is impossible to learn more global features. This makes the performance of the VT series models unable to improve after data augmentation significantly. However, the performance of BotNet, a hybrid model of CNN and VT, is significantly improved after data augmentation. This is because the BotNet network only replaced three Bottlenecks with MHSA. The BotNet network is essentially more inclined to the feature extraction method of CNN.

\section{Conclusion and Future Work\label{sec: Conclusion and Future Work}}

The classification of small EM datasets are very challenging in computer vision tasks, which has attracted the attention of many researchers. Due to the development of deep learning, image classification of small datasets is developing rapidly. This article uses 17 CNN models, 3 VT models, and a hybrid CNN and VT model to test model performance. We have performed several experiments, including direct classification of each model, classification tasks after data augmentation, and imbalanced training tasks on some representative models. The experimental results prove that the Xception network is suitable for this kind of task.
The ViT models take the least time for training. Therefore, the ViT model is suitable for large-scale data training. The ShuffleNet-V2 network has the least number of parameters, although its classification performance is average. Therefore, ShuffleNet-V2 is more suitable for occasions where high classification performance is not necessary and limited storage space.

This study provides an analysis table of the differences between the 18 models. This result can help related research on feature fusion quickly find models with significant differences and improve model performance. In addition, this study finds for the first time that the data augmentation method of geometric deformation is extremely limited or even ineffective in improving the performance of VT series models. This study and conclusion can provide relevant researchers with a conclusion with sufficient experimental support. Our research and conclusions reduce their workload in selecting experimental augmentation methods to a certain extent. This has a significant reference value.

Although the augmentation method of geometric deformation is effective for the performance improvement of CNNs, it does not help much for the performance improvement of VTs. We can improve the VT networks performance by studying new data augmentation methods in future work.

\bibliography{Zhao}

\clearpage
\bibliographystyle{unsrt}
  \onecolumn
\section*{Appendix}

\begin{center}
\renewcommand\arraystretch{1.3} 
\begin{longtable}{lllllll}
\hline
\multicolumn{2}{l}{Model}             & Change (up) &  & \multicolumn{2}{l}{Model}             & Accuracy \\ \hline
ResNet101         & VGG11             & 9.52\%   &  & Googlenet         & Xception          & 46.03\%  \\
InceptionResnetV1 & ResNet18          & 7.94\%   &  & Inception-V3      & Xception          & 44.76\%  \\
Inception-V3      & Shufflenet-V2     & 7.62\%   &  & ResNet50          & Xception          & 44.76\%  \\
Shufflenet-V2     & VGG11             & 7.62\%   &  & Deit              & Xception          & 44.44\%  \\
Deit              & VGG11             & 7.30\%   &  & Densenet161       & Xception          & 44.13\%  \\
Inception-V3      & VGG11             & 7.30\%   &  & VGG11             & Xception          & 44.13\%  \\
ResNet18          & ResNet50          & 7.30\%   &  & Densenet121       & Xception          & 43.81\%  \\
ResNet34          & ResNet50          & 7.30\%   &  & Mobilenet-V2      & Xception          & 43.81\%  \\
ResNet34          & VGG11             & 7.30\%   &  & ResNet34          & ResNet50          & 43.81\%  \\
ResNet101         & Shufflenet-V2     & 7.30\%   &  & ResNet34          & VGG11             & 43.81\%  \\
Googlenet         & Mobilenet-V2      & 7.30\%   &  & Densenet121       & ResNet34          & 43.49\%  \\
Alexnet           & T2T-VIT          & 6.98\%   &  & Googlenet         & ResNet34          & 43.49\%  \\
Deit              & Mobilenet-V2      & 6.98\%   &  & InceptionResnetV1 & Xception          & 43.49\%  \\
Deit              & vit-5             & 6.98\%   &  & Mobilenet-V2      & ResNet34          & 43.49\%  \\
Densenet121       & Googlenet         & 6.98\%   &  & ResNet18          & Xception          & 43.49\%  \\
Densenet121       & Mobilenet-V2      & 6.98\%   &  & ResNet34          & Xception          & 43.17\%  \\
Densenet121       & ResNet34          & 6.98\%   &  & InceptionResnetV1 & ResNet34          & 42.86\%  \\
Googlenet         & ResNet34          & 6.98\%   &  & Inception-V3      & ResNet34          & 42.86\%  \\
Googlenet         & VGG11             & 6.98\%   &  & vit-5             & Xception          & 42.86\%  \\
Inception-V3      & vit-5             & 6.98\%   &  & Googlenet         & Mobilenet-V2      & 42.54\%  \\
Mobilenet-V2      & ResNet18          & 6.98\%   &  & ResNet101         & Xception          & 42.54\%  \\
Mobilenet-V2      & ResNet34          & 6.98\%   &  & Densenet121       & Googlenet         & 42.22\%  \\
Alexnet           & Shufflenet-V2     & 6.67\%   &  & Googlenet         & VGG11             & 42.22\%  \\
Densenet121       & ResNet18          & 6.67\%   &  & Densenet161       & Googlenet         & 41.90\%  \\
ResNet18          & VGG11             & 6.67\%   &  & Densenet161       & ResNet34          & 41.90\%  \\
Alexnet           & ResNet101         & 6.67\%   &  & Deit              & ResNet34          & 41.59\%  \\
Densenet121       & Shufflenet-V2     & 6.67\%   &  & Googlenet         & InceptionResnetV1 & 41.59\%  \\
Densenet161       & Googlenet         & 6.67\%   &  & InceptionResnetV1 & ResNet18          & 41.59\%  \\
Deit              & Inception-V3      & 6.35\%   &  & ResNet34          & vit-5             & 41.59\%  \\
Deit              & Shufflenet-V2     & 6.35\%   &  & Deit              & Mobilenet-V2      & 41.27\%  \\
Densenet161       & Mobilenet-V2      & 6.35\%   &  & Densenet121       & Mobilenet-V2      & 41.27\%  \\
Googlenet         & InceptionResnetV1 & 6.35\%   &  & Googlenet         & ResNet50          & 41.27\%  \\
InceptionResnetV1 & ResNet34          & 6.35\%   &  & Mobilenet-V2      & ResNet18          & 41.27\%  \\
Inception-V3      & ResNet18          & 6.35\%   &  & Shufflenet-V2     & Xception          & 41.27\%  \\
Inception-V3      & ResNet34          & 6.35\%   &  & T2T-VIT          & Xception          & 41.27\%  \\
Mobilenet-V2      & vit-5             & 6.35\%   &  & Alexnet           & Xception          & 40.95\%  \\
ResNet50          & Shufflenet-V2     & 6.35\%   &  & Botnet            & Xception          & 40.63\%  \\
Deit              & ResNet18          & 6.03\%   &  & Densenet161       & Mobilenet-V2      & 40.63\%  \\
InceptionResnetV1 & Mobilenet-V2      & 6.03\%   &  & Googlenet         & vit-5             & 40.63\%  \\
Inception-V3      & Mobilenet-V2      & 6.03\%   &  & Mobilenet-V2      & vit-5             & 40.63\%  \\
ResNet50          & vit-5             & 6.03\%   &  & ResNet18          & ResNet50          & 40.63\%  \\
\hline
\multicolumn{2}{l}{Model}             & Change (up) &  & \multicolumn{2}{l}{Model}             & Accuracy \\ \hline
VGG11             & vit-5             & 6.03\%   &  & Googlenet         & Inception-V3      & 40.32\%  \\
Deit              & InceptionResnetV1 & 6.03\%   &  & InceptionResnetV1 & Mobilenet-V2      & 40.32\%  \\
Densenet161       & Inception-V3      & 6.03\%   &  & Inception-V3      & Mobilenet-V2      & 40.32\%  \\
Googlenet         & ResNet50          & 6.03\%   &  & ResNet18          & ResNet34          & 40.32\%  \\
Botnet            & VGG11             & 5.71\%   &  & Densenet121       & ResNet18          & 40.00\%  \\
Densenet161       & InceptionResnetV1 & 5.71\%   &  & Googlenet         & ResNet18          & 40.00\%  \\
Googlenet         & Xception          & 5.71\%   &  & Mobilenet-V2      & ResNet50          & 40.00\%  \\
InceptionResnetV1 & Shufflenet-V2     & 5.71\%   &  & ResNet18          & VGG11             & 40.00\%  \\
Mobilenet-V2      & ResNet50          & 5.71\%   &  & Alexnet           & ResNet34          & 39.68\%  \\
Alexnet           & Botnet            & 5.40\%   &  & Deit              & InceptionResnetV1 & 39.68\%  \\
Deit              & Densenet121       & 5.40\%   &  & Densenet121       & Shufflenet-V2     & 39.68\%  \\
Densenet161       & ResNet34          & 5.40\%   &  & Densenet161       & Inception-V3      & 39.68\%  \\
ResNet101         & vit-5             & 5.40\%   &  & Inception-V3      & ResNet18          & 39.68\%  \\
Densenet121       & InceptionResnetV1 & 5.40\%   &  & ResNet50          & Shufflenet-V2     & 39.68\%  \\
Densenet161       & ResNet18          & 5.40\%   &  & Deit              & Googlenet         & 39.37\%  \\
Googlenet         & vit-5             & 5.40\%   &  & Deit              & ResNet18          & 39.37\%  \\
InceptionResnetV1 & ResNet50          & 5.40\%   &  & Densenet161       & InceptionResnetV1 & 39.37\%  \\
InceptionResnetV1 & vit-5             & 5.40\%   &  & InceptionResnetV1 & Shufflenet-V2     & 39.37\%  \\
Botnet            & Shufflenet-V2     & 5.08\%   &  & ResNet50          & vit-5             & 39.37\%  \\
Deit              & ResNet34          & 5.08\%   &  & Alexnet           & Googlenet         & 39.05\%  \\
ResNet34          & vit-5             & 5.08\%   &  & Densenet121       & InceptionResnetV1 & 39.05\%  \\
Alexnet           & VGG11             & 5.08\%   &  & Densenet161       & ResNet18          & 39.05\%  \\
Deit              & Densenet161       & 5.08\%   &  & Googlenet         & Shufflenet-V2     & 39.05\%  \\
Densenet121       & Inception-V3      & 5.08\%   &  & InceptionResnetV1 & ResNet50          & 39.05\%  \\
Densenet161       & ResNet50          & 5.08\%   &  & InceptionResnetV1 & vit-5             & 39.05\%  \\
Googlenet         & Inception-V3      & 5.08\%   &  & Mobilenet-V2      & ResNet101         & 39.05\%  \\
InceptionResnetV1 & Inception-V3      & 5.08\%   &  & ResNet34          & ResNet101         & 39.05\%  \\
InceptionResnetV1 & VGG11             & 5.08\%   &  & Deit              & Densenet161       & 38.73\%  \\
T2T-VIT           & VGG11             & 5.08\%   &  & Deit              & vit-5             & 38.73\%  \\
Botnet            & T2T-VIT          & 4.76\%   &  & Densenet161       & ResNet50          & 38.73\%  \\
Deit              & ResNet50          & 4.76\%   &  & Googlenet         & ResNet101         & 38.73\%  \\
Deit              & ResNet101         & 4.76\%   &  & InceptionResnetV1 & Inception-V3      & 38.73\%  \\
Densenet121       & Densenet161       & 4.76\%   &  & InceptionResnetV1 & VGG11             & 38.73\%  \\
Densenet121       & vit-5             & 4.76\%   &  & Inception-V3      & Shufflenet-V2     & 38.73\%  \\
Densenet161       & vit-5             & 4.76\%   &  & Inception-V3      & vit-5             & 38.73\%  \\
Googlenet         & ResNet18          & 4.76\%   &  & ResNet34          & Shufflenet-V2     & 38.73\%  \\
Mobilenet-V2      & ResNet101         & 4.76\%   &  & Botnet            & ResNet34          & 38.41\%  \\
Alexnet           & Inception-V3      & 4.44\%   &  & Deit              & Densenet121       & 38.41\%  \\
Inception-V3      & ResNet50          & 4.44\%   &  & Deit              & VGG11             & 38.41\%  \\
Inception-V3      & Xception          & 4.44\%   &  & Densenet121       & Densenet161       & 38.41\%  \\
ResNet50          & Xception          & 4.44\%   &  & Densenet161       & vit-5             & 38.41\%  \\
Deit              & Googlenet         & 4.13\%   &  & Inception-V3      & VGG11             & 38.41\%  \\
\hline
\multicolumn{2}{l}{Model}             & Change (up) &  & \multicolumn{2}{l}{Model}             & Accuracy \\ \hline
Deit              & Xception          & 4.13\%   &  & Deit              & ResNet50          & 38.10\%  \\
Densenet161       & Shufflenet-V2     & 4.13\%   &  & Densenet121       & Inception-V3      & 38.10\%  \\
Densenet161       & VGG11             & 4.13\%   &  & Googlenet         & T2T-VIT           & 38.10\%  \\
InceptionResnetV1 & ResNet101         & 4.13\%   &  & Mobilenet-V2      & VGG11             & 38.10\%  \\
Inception-V3      & ResNet101         & 4.13\%   &  & Botnet            & Googlenet         & 37.78\%  \\
ResNet50          & ResNet101         & 4.13\%   &  & Densenet121       & vit-5             & 37.78\%  \\
Alexnet           & ResNet18          & 3.81\%   &  & Densenet161       & Shufflenet-V2     & 37.78\%  \\
Alexnet           & ResNet50          & 3.81\%   &  & Densenet161       & VGG11             & 37.78\%  \\
Densenet121       & ResNet101         & 3.81\%   &  & InceptionResnetV1 & ResNet101         & 37.78\%  \\
ResNet18          & ResNet34          & 3.81\%   &  & Inception-V3      & ResNet50          & 37.78\%  \\
ResNet50          & VGG11             & 3.81\%   &  & Mobilenet-V2      & Shufflenet-V2     & 37.78\%  \\
Shufflenet-V2     & T2T-VIT          & 3.81\%   &  & ResNet34          & T2T-VIT           & 37.78\%  \\
Shufflenet-V2     & vit-5             & 3.81\%   &  & VGG11             & vit-5             & 37.78\%  \\
Alexnet           & Googlenet         & 3.81\%   &  & Deit              & Inception-V3      & 37.46\%  \\
Botnet            & Deit              & 3.81\%   &  & Deit              & Shufflenet-V2     & 37.46\%  \\
Densenet161       & ResNet101         & 3.81\%   &  & Densenet161       & ResNet101         & 37.46\%  \\
Densenet161       & Xception          & 3.81\%   &  & ResNet50          & ResNet101         & 37.46\%  \\
Googlenet         & Shufflenet-V2     & 3.81\%   &  & ResNet101         & VGG11             & 37.46\%  \\
Mobilenet-V2      & VGG11             & 3.81\%   &  & Alexnet           & Mobilenet-V2      & 37.14\%  \\
VGG11             & Xception          & 3.81\%   &  & Alexnet           & ResNet18          & 37.14\%  \\
Botnet            & ResNet101         & 3.49\%   &  & Alexnet           & ResNet50          & 37.14\%  \\
Botnet            & vit-5             & 3.49\%   &  & Mobilenet-V2      & T2T-VIT           & 37.14\%  \\
Densenet121       & ResNet50          & 3.49\%   &  & ResNet50          & VGG11             & 37.14\%  \\
Densenet121       & VGG11             & 3.49\%   &  & ResNet101         & vit-5             & 37.14\%  \\
Densenet121       & Xception          & 3.49\%   &  & Densenet121       & ResNet50          & 36.83\%  \\
Googlenet         & ResNet101         & 3.49\%   &  & Densenet121       & ResNet101         & 36.83\%  \\
Mobilenet-V2      & Shufflenet-V2     & 3.49\%   &  & ResNet18          & ResNet101         & 36.83\%  \\
Mobilenet-V2      & Xception          & 3.49\%   &  & ResNet18          & vit-5             & 36.83\%  \\
ResNet18          & ResNet101         & 3.49\%   &  & Densenet121       & VGG11             & 36.51\%  \\
ResNet18          & vit-5             & 3.49\%   &  & Densenet161       & T2T-VIT           & 36.51\%  \\
Alexnet           & Deit              & 3.17\%   &  & ResNet18          & T2T-VIT           & 36.51\%  \\
Alexnet           & ResNet34          & 3.17\%   &  & Botnet            & Densenet161       & 36.19\%  \\
Deit              & T2T-VIT           & 3.17\%   &  & Botnet            & Mobilenet-V2      & 36.19\%  \\
InceptionResnetV1 & Xception          & 3.17\%   &  & Botnet            & ResNet18          & 36.19\%  \\
ResNet18          & T2T-VIT           & 3.17\%   &  & ResNet18          & Shufflenet-V2     & 36.19\%  \\
ResNet18          & Xception          & 3.17\%   &  & Botnet            & Densenet121       & 35.87\%  \\
T2T-VIT          & vit-5             & 3.17\%   &  & Deit              & ResNet101         & 35.87\%  \\
Alexnet           & Mobilenet-V2      & 2.86\%   &  & Alexnet           & Densenet161       & 35.56\%  \\
Alexnet           & vit-5             & 2.86\%   &  & Alexnet           & InceptionResnetV1 & 35.56\%  \\
Botnet            & Densenet121       & 2.86\%   &  & Alexnet           & Inception-V3      & 35.56\%  \\
Botnet            & Inception-V3      & 2.86\%   &  & Botnet            & InceptionResnetV1 & 35.56\%  \\
Botnet            & ResNet18          & 2.86\%   &  & Botnet            & ResNet50          & 35.56\%  \\
\hline
\multicolumn{2}{l}{Model}             & Change (up) &  & \multicolumn{2}{l}{Model}             & Accuracy \\ \hline
Densenet161       & T2T-VIT           & 2.86\%   &  & Densenet121       & T2T-VIT           & 35.56\%  \\
Googlenet         & T2T-VIT           & 2.86\%   &  & InceptionResnetV1 & T2T-VIT           & 35.56\%  \\
Mobilenet-V2      & T2T-VIT           & 2.86\%   &  & ResNet50          & T2T-VIT           & 35.56\%  \\
ResNet18          & Shufflenet-V2     & 2.86\%   &  & Shufflenet-V2     & vit-5             & 35.56\%  \\
ResNet34          & Xception          & 2.86\%   &  & Alexnet           & Densenet121       & 35.24\%  \\
ResNet101         & T2T-VIT           & 2.86\%   &  & Botnet            & vit-5             & 35.24\%  \\
Densenet121       & T2T-VIT           & 2.54\%   &  & Inception-V3      & ResNet101         & 35.24\%  \\
ResNet34          & ResNet101         & 2.54\%   &  & ResNet101         & Shufflenet-V2     & 35.24\%  \\
Botnet            & Densenet161       & 2.54\%   &  & Shufflenet-V2     & VGG11             & 35.24\%  \\
Botnet            & Googlenet         & 2.54\%   &  & Botnet            & Deit              & 34.92\%  \\
vit-5             & Xception          & 2.54\%   &  & T2T-VIT           & vit-5             & 34.92\%  \\
Alexnet           & Densenet121       & 2.22\%   &  & Alexnet           & ResNet101         & 34.60\%  \\
Botnet            & ResNet50          & 2.22\%   &  & Alexnet           & vit-5             & 34.60\%  \\
Inception-V3      & T2T-VIT           & 2.22\%   &  & Alexnet           & Deit              & 34.29\%  \\
ResNet34          & Shufflenet-V2     & 2.22\%   &  & Deit              & T2T-VIT           & 34.29\%  \\
ResNet50          & T2T-VIT           & 2.22\%   &  & Alexnet           & Shufflenet-V2     & 33.97\%  \\
ResNet101         & Xception          & 2.22\%   &  & Botnet            & Inception-V3      & 33.97\%  \\
Alexnet           & Densenet161       & 1.90\%   &  & Botnet            & VGG11             & 33.33\%  \\
Alexnet           & InceptionResnetV1 & 1.90\%   &  & Inception-V3      & T2T-VIT           & 33.33\%  \\
Botnet            & InceptionResnetV1 & 1.90\%   &  & Alexnet           & VGG11             & 32.70\%  \\
Botnet            & Mobilenet-V2      & 1.90\%   &  & T2T-VIT           & VGG11             & 32.70\%  \\
Botnet            & ResNet34          & 1.90\%   &  & Botnet            & Shufflenet-V2     & 32.38\%  \\
InceptionResnetV1 & T2T-VIT           & 1.90\%   &  & Alexnet           & T2T-VIT           & 31.43\%  \\
ResNet34          & T2T-VIT           & 1.27\%   &  & Botnet            & ResNet101         & 31.43\%  \\
Shufflenet-V2     & Xception          & 0.95\%   &  & Shufflenet-V2     & T2T-VIT           & 31.11\%  \\
T2T-VIT           & Xception          & 0.95\%   &  & Alexnet           & Botnet            & 30.79\%  \\
Alexnet           & Xception          & 0.63\%   &  & ResNet101         & T2T-VIT           & 30.79\%  \\
Botnet            & Xception          & 0.32\%   &  & Botnet            & T2T-VIT           & 30.16\% \\
\hline
\end{longtable}
\end{center}

\end{document}